\documentclass{article}
\PassOptionsToPackage{numbers,compress}{natbib}        
\usepackage[preprint]{neurips_2026}

\usepackage[utf8]{inputenc}
\usepackage[T1]{fontenc}
\usepackage{hyperref}
\usepackage{url}
\usepackage{booktabs}
\usepackage{amsfonts}
\usepackage{nicefrac}
\usepackage{microtype}
\usepackage{xcolor}

\usepackage{graphicx}
\usepackage{enumitem}
\usepackage{amsmath}
\usepackage{amssymb}
\usepackage{subcaption}
\usepackage{multirow}
\usepackage{tabularx}
\usepackage{placeins}
\usepackage{pifont}
\usepackage{tikz}
\usetikzlibrary{calc}
\usepackage{pgfplots}
\usepgfplotslibrary{statistics}
\pgfplotsset{compat=1.18}
\usepgfplotslibrary{fillbetween}
\usepackage[ruled,linesnumbered]{algorithm2e}
\usepackage[capitalise,noabbrev]{cleveref}
\hypersetup{colorlinks=true, linkcolor=blue!60!black, citecolor=green!40!black, urlcolor=magenta!60!black}

\usepackage{listings}
\usetikzlibrary{arrows.meta, calc, angles, quotes, decorations.pathreplacing}
\newcommand{\githubicon}{\raisebox{-0.12em}{\tikz[x=0.058em,y=0.058em]\fill[black] (8.00,16.00)..controls(3.58,16.00)and(0.00,12.42)..(0.00,8.00)..controls(0.00,4.46)and(2.29,1.47)..(5.47,0.41)..controls(5.87,0.34)and(6.02,0.58)..(6.02,0.79)..controls(6.02,0.98)and(6.01,1.61)..(6.01,2.28)..controls(4.00,1.91)and(3.48,2.77)..(3.32,3.22)..controls(3.23,3.45)and(2.84,4.16)..(2.50,4.35)..controls(2.22,4.50)and(1.82,4.87)..(2.49,4.88)..controls(3.12,4.89)and(3.57,4.30)..(3.72,4.06)..controls(4.44,2.85)and(5.59,3.19)..(6.05,3.40)..controls(6.12,3.92)and(6.33,4.27)..(6.56,4.47)..controls(4.78,4.67)and(2.92,5.36)..(2.92,8.42)..controls(2.92,9.29)and(3.23,10.01)..(3.74,10.57)..controls(3.66,10.77)and(3.38,11.59)..(3.82,12.69)..controls(3.82,12.69)and(4.49,12.90)..(6.02,11.87)..controls(6.66,12.05)and(7.34,12.14)..(8.02,12.14)..controls(8.70,12.14)and(9.38,12.05)..(10.02,11.87)..controls(11.55,12.91)and(12.22,12.69)..(12.22,12.69)..controls(12.66,11.59)and(12.38,10.77)..(12.30,10.57)..controls(12.81,10.01)and(13.12,9.30)..(13.12,8.42)..controls(13.12,5.35)and(11.25,4.67)..(9.47,4.47)..controls(9.76,4.22)and(10.01,3.74)..(10.01,2.99)..controls(10.01,1.92)and(10.00,1.06)..(10.00,0.79)..controls(10.00,0.58)and(10.15,0.33)..(10.55,0.41)--(10.94,0.55)--(11.32,0.72)--(11.70,0.90)--(12.06,1.10)--(12.41,1.32)--(12.75,1.56)--(13.08,1.82)--(13.39,2.09)--(13.69,2.38)--(13.98,2.68)--(14.24,3.00)--(14.50,3.33)--(14.73,3.67)--(14.94,4.03)--(15.14,4.39)--(15.32,4.77)--(15.48,5.15)--(15.61,5.55)--(15.73,5.94)--(15.83,6.35)--(15.90,6.76)--(15.96,7.17)--(15.99,7.58)--(16.00,8.00)..controls(16.00,12.42)and(12.42,16.00)..(8.00,16.00)--cycle;}}
\usepackage{adjustbox}

\colorlet{myblue}{blue!70}
\colorlet{myorange}{orange!90!black}
\colorlet{mygreen}{green!60!black!80}
\colorlet{mypurple}{violet!70}

\newcommand{\hij}{\mathbf{h}_{ij}}

\newcommand{\hhij}{\hat{\mathbf{h}}_{ij}}
\newcommand{\hhkl}{\hat{\mathbf{h}}_{k\ell}}
\newcommand{\mubar}{\boldsymbol{\bar{\mu}}}
\newcommand{\hhklmax}{\hat{\mathbf{h}}_{k^\star\ell^\star}}
\newcommand{\xbf}{\mathbf{x}}

\newcommand{\vbf}{\mathbf{v}}
\newcommand{\sbf}{\mathbf{s}}
\newcommand{\ubf}{\mathbf{u}}

\title{TokenRepel: Generating Diverse Image Sets\\Without Compromising Quality}

\definecolor{alphaColor}{HTML}{FF9100}
\definecolor{betaColor}{HTML}{00D5FF}
\definecolor{gammaColor}{HTML}{00B050}
\newcommand{\instalpha}{{\color{alphaColor}{\alpha}}}
\newcommand{\instbeta}{{\color{betaColor}{\beta}}}
\newcommand{\instgamma}{{\color{gammaColor}{\gamma}}}

\author{%
  Qingtao Yu$^{\instalpha}$ \quad Zhen Wang$^{\instalpha\instbeta}$ \quad Ning Lai$^{\instalpha\instgamma}$ \quad
  Boyang Hu$^{\instalpha}$ \quad Hongdong Li$^{\instalpha}$ \quad Dylan Campbell$^{\instalpha}$ \\
  $^{\instalpha}$The Australian National University \quad
  $^{\instbeta}$Homenet Research \quad
  $^{\instgamma}$CarryAI \\
  {\small\texttt{\{terry.yu,\,zhen.wang,\,ning.lai,\,boyang.hu,\,hongdong.li,\,dylan.campbell\}@anu.edu.au}}
}

\begin{document}

\maketitle

\begin{center}
    \captionsetup{type=figure}
    \makebox[\linewidth][c]{\begin{tikzpicture}[font=\fontfamily{phv}\selectfont, inner sep=0pt, outer sep=0pt]
    \def\nrows{3}
    \pgfmathsetlengthmacro{\figw}{1.12\linewidth}
    \pgfmathsetlengthmacro{\imgsep}{1pt}
    \pgfmathsetlengthmacro{\groupsep}{4pt}
    \pgfmathsetlengthmacro{\pad}{3pt}
    \pgfmathsetlengthmacro{\imgw}{(\figw - 6*\imgsep - 4*\pad - \groupsep)/8}
    \pgfmathsetlengthmacro{\groupw}{4*\imgw + 3*\imgsep}
    \pgfmathsetlengthmacro{\xbase}{\pad}
    \pgfmathsetlengthmacro{\xours}{\groupw + 3*\pad + \groupsep}
    \pgfmathsetlengthmacro{\rowh}{\imgw + 15pt}

    \pgfmathsetlengthmacro{\boxtop}{16pt}
    \pgfmathsetlengthmacro{\boxbot}{-(\nrows - 1)*\rowh - \imgw - 16pt}
    \fill[blue!8, rounded corners=4pt] (0, \boxtop) rectangle ({\groupw + 2*\pad}, \boxbot);
    \fill[yellow!25, rounded corners=4pt] ({\xours - \pad}, \boxtop) rectangle (\figw, \boxbot);

    \newcommand{\teaserrow}[5]{
        \pgfmathsetlengthmacro{\y}{-#1*\rowh}
        \foreach \i in {0,...,3} {
            \node[anchor=north west] at ({\xbase + \i*(\imgw + \imgsep)}, \y)
                {\includegraphics[width=\imgw]{figures_arxiv/teaser/images/#3_\i.jpg}};
            \node[anchor=north west] at ({\xours + \i*(\imgw + \imgsep)}, \y)
                {\includegraphics[width=\imgw]{figures_arxiv/teaser/images/#4_\i.jpg}};
        }
        \node[anchor=north, fill=white, rounded corners=2pt, inner xsep=4pt, inner ysep=1.5pt,
              font=\fontfamily{phv}\selectfont\scriptsize]
            at ({\figw/2}, {\y - \imgw - 3pt}) {\textbf{#2}: \textit{``#5''}};
    }

    \node[anchor=south, font=\fontfamily{phv}\selectfont\small]
        at ({\xbase + \groupw/2}, 3pt) {\textbf{Baseline}};
    \node[anchor=south, font=\fontfamily{phv}\selectfont\small]
        at ({\xours + \groupw/2}, 3pt) {\textbf{With Ours}};

    \teaserrow{0}{FLUX.1-dev}{flux_base_p014}{flux_ours_p014}{A ballerina dancing in an empty train station.}
    \teaserrow{1}{FLUX.2-Klein-9B}{flux2k_base_p000}{flux2k_ours_p000}{A girl with pink hair and a face tattoo.}
    \teaserrow{2}{SDXL}{sdxl_base_p032}{sdxl_ours_p032}{A hot air balloon over a field of sunflowers.}
\end{tikzpicture}
}
    \vspace{-10pt}
    \caption{Qualitative results of TokenRepel on three representative text-to-image models.}
    \label{fig:teaser}
\end{center}

\begin{abstract}
We present an inference-time approach that enhances the visual diversity of image sets generated by text-to-image models,
without compromising the average image quality.
Our method leverages the intrinsic geometry of hidden representations in pretrained models
to diversify images generated from the same prompt.
First, we introduce a token repulsion mechanism that increases the angle between feature tokens across samples.
It takes a negative Riemannian gradient step on a von Mises--Fisher mixture distribution of tokens on the unit hypersphere.
Second, to preserve the quality of each output image,
we propose a token density preservation mechanism that projects the repulsion-induced prediction change onto a half-space aligned with the model's score direction.
This encourages the updates to remain within the high-density regions learned by the generative model.
We further extend this mechanism to the output of each intermediate layer,
enabling finer-grained iterative guidance.
We evaluate our method on three types of pretrained text-to-image models.
Experiments show substantial gains in visual diversity under identical prompts,
while maintaining a comparable level of image fidelity.
Code is available at \href{https://github.com/TerrysLearning/TokenRepel_Official}{\githubicon\,\nolinkurl{github.com/TerrysLearning/TokenRepel_Official}}.

\end{abstract}

\section{Introduction}
Recent advances in text-to-image (T2I) generation have led to unprecedented levels of visual quality \cite{flux, sd, sd3, sdxl, qwenimage, infinity}.
However, a challenge persists in the inherent trade-off between image quality and sample diversity \cite{cfg, diversegrpo, grpo_guard, gass, cads}.
Diversity, defined as a T2I model's ability to generate varied outputs for a fixed condition, is a critical component of generative modeling.
Enhancing diversity can foster creativity and provide designers with a broader range of visual choices \cite{cads, dreampose}, 
support data augmentation \cite{gendataargu} and test-time scaling \cite{tts_ma, tts_tut}, and increase the representation of ethnicities and genders underrepresented in the data \cite{minority}.
Although diffusion and flow-matching models are capable of broad mode coverage, 
they may suffer from ``mode gravity'' during inference \cite{closerlook}. 
In these scenarios, models tend to generate highly similar outputs regardless of the random initial noise,
focusing on a narrow subset of the data distribution to generate only the highest confidence samples.

Existing inference-time methods diversify samples by changing prompts or text conditioning \cite{prompt_expansion,cads}, 
optimizing initial noise \cite{noise_opt_diverse,diverse_contrat_noise}, or repelling image latents \cite{sparke,particle_g}.
More recent methods intervene in intermediate features, which carry richer information about the generated content \cite{dsm,contextual_repulsion}.
However, they treat each feature representation as a whole rather than explicitly modeling interactions between individual tokens.
Token-level repulsion can act more selectively, pushing highly similar cross-image tokens apart while leaving unrelated tokens untouched.
For example, a token representing a dog's paw should be repelled from a similar paw token in another image, but not from a sky token.
NegToMe \cite{negtome} follows this token-level approach, but its linear extrapolation has undesirable mathematical properties, limiting its effectiveness.
It also lacks an explicit fidelity-preserving mechanism, which can introduce artifacts.

To address this problem, we present TokenRepel, a training-free algorithm for generating diverse image sets while preserving fidelity.
To model cross-image token interactions, we represent the normalized tokens from other images as a von Mises--Fisher (vMF) mixture on the unit hypersphere.
We then update each token with a negative Riemannian gradient step on this density, increasing the angular separation between similar tokens.
To prevent these updates from degrading fidelity, we introduce a token density preservation mechanism.
It projects the repulsion-induced prediction change onto a half-space aligned with the model's score direction, 
encouraging tokens to stay in high-density regions.
We further extend this mechanism to intermediate layers, allowing for more frequent and fine-grained corrections.
Together, token repulsion and token density preservation increase the diversity of generated images while largely maintaining their fidelity.

Our method applies to different architectures without requiring external models or additional training.
We evaluate our approach on three representative T2I models: FLUX.1-dev \cite{flux} (flow-matching), SDXL \cite{sdxl} (diffusion), and FLUX.2-Klein \cite{flux2klein} (few-step distilled flow-matching). 
We achieve a more favorable diversity--quality trade-off than existing inference-time methods. 
Our approach is simple and effective, with moderate computational overhead.

\section{Related Work}

Although diffusion and flow-matching models improve both diversity and fidelity over GANs \cite{gan, diffusionbeat, flow_gen}, 
further enhancing the sampling diversity remains a critical challenge \cite{closerlook}. 
This issue is further compounded by recent trends in model optimization,
as both post-training \cite{grpo_guard, diversegrpo, imagerefl} and distillation \cite{distill_lose_diverse} have been shown to diminish diversity.
To mitigate this, \citet{prompt_expansion} use specialized LLMs to generate diverse descriptions.
\citet{noise_opt_diverse} repeatedly generate image batches and update their initial noise using diversity and quality objectives.
GASS decomposes image embeddings into orthogonal axes for independent prompt-dependent and prompt-independent variation \cite{gass}. 
Group Inference generates a large candidate pool and selects a subset by jointly considering image quality and group diversity \cite{group_inference}.
These approaches may require external models, additional decoding, or a much larger candidate pool. 

Other inference-time interventions have also been proposed.
\citet{minority} introduced guidance that moves samples beyond high-density regions.
Contrastive Noise Optimization (CNO) optimizes batches of initial noise latents with a contrastive loss on their Tweedie estimates \cite{cno}.
CADS perturbs text embeddings early in sampling according to a schedule \cite{cads}.
SPARKE and Particle Guidance (PG) optimize batch latents with dissimilarity objectives based on inverse R\'enyi kernel entropy and pairwise RBF kernels, respectively \cite{sparke, particle_g}.
Shielded Diffusion applies sparse repulsion to nearby Tweedie estimates \cite{shielded_diffusion};
DSM disperses intermediate features through feature self-guidance \cite{dsm}; 
and Contextual Repulsion repels contextual embeddings in multimodal attention \cite{contextual_repulsion}.
These methods generally do not model fine-grained interactions between image feature tokens or explicitly correct diversity-promoting updates to preserve fidelity.
NegToMe linearly extrapolates attention tokens away from a reference image within the batch \cite{negtome}, but this update can push less similar tokens farther apart.
Motion Modes~\cite{motion_modes} also uses repulsion to enhance diversity in video generation.
It guides each new object motion away from previously generated motion fields.
TokenRepel instead repels similar hidden tokens across images using angular feature geometry, then corrects the induced predictions toward a high-density half-space to preserve fidelity.

\begin{figure}[t]
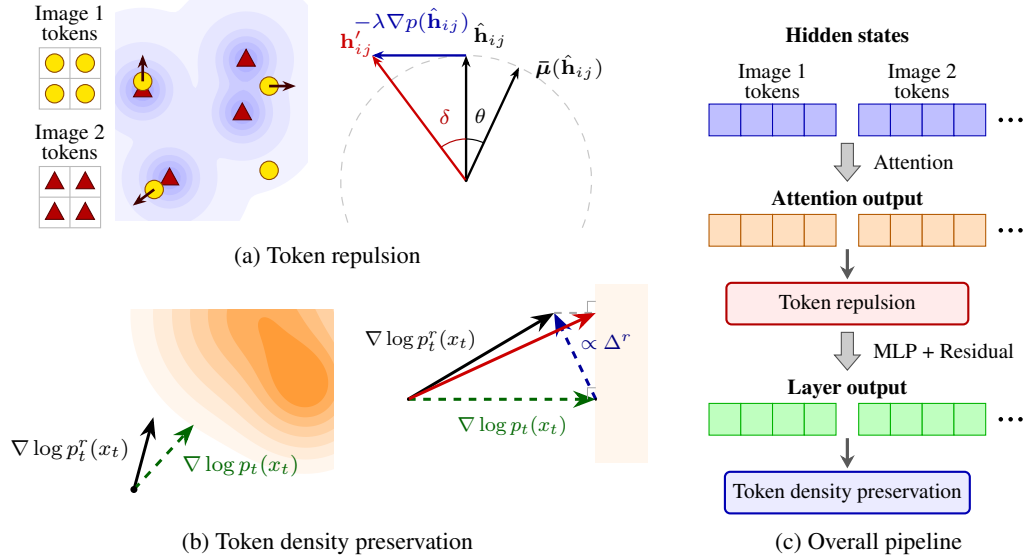

    \centering
    \begin{minipage}[b]{0.655\textwidth}
        \begin{subfigure}[b]{\linewidth}
            \centering
            \adjustbox{max width=0.49\linewidth, max totalheight=0.333\linewidth}{\resizebox{0.49\linewidth}{!}{\begin{tikzpicture}[
    scale=0.5,
    >=Stealth,
    push_arrow/.style    ={-{Stealth[length=5pt]}, line width=1.0pt, red!25!black},
    push_long/.style     ={-{Stealth[length=6pt]}, line width=1.2pt, red!25!black},
    sec3_vec/.style      ={-{Stealth[length=8pt]}, thick, black},
    sec3_redvec/.style   ={-{Stealth[length=8pt]}, thick, red!80!black},
    sec3_pushvec/.style  ={-{Stealth[length=8pt]}, thick, blue!65!black},
    sec3_rotarrow/.style ={-{Stealth[length=7pt]}, thick, red!65!black},
    callout/.style       ={dashed, gray!150, thick},
]

\tikzset{
    yellowtok/.style ={fill=yellow!88!orange, draw=orange!55!black, line width=0.4pt},
    pics/circletok/.style  ={code={\path[yellowtok] (0,0) circle (0.12);}},
    pics/triangletok/.style={code={\path[yellowtok] (-0.12,-0.09) -- (0.12,-0.09) -- (0,0.13) -- cycle;}},
    pics/squaretok/.style  ={code={\path[yellowtok] (-0.11,-0.11) rectangle (0.11,0.11);}},
    pics/hearttok/.style   ={code={
        \path[yellowtok] (0,0.03)
            ..controls (0.12,0.19) and (0.22,0.03) .. (0,-0.13)
            ..controls (-0.22,0.03) and (-0.12,0.19) .. cycle;
    }},
    redtok/.style    ={fill=red!70!black, draw=red!35!black, line width=0.4pt},
    pics/redtriangletok/.style={code={\path[redtok] (-0.12,-0.09) -- (0.12,-0.09) -- (0,0.13) -- cycle;}},
    pics/crosstok/.style   ={code={
        \draw[black, line width=1.8pt, line cap=round] (-0.09,-0.09) -- (0.09,0.09);
        \draw[black, line width=1.8pt, line cap=round] (-0.09, 0.09) -- (0.09,-0.09);
        \draw[red!75!black, line width=1.0pt, line cap=round] (-0.09,-0.09) -- (0.09,0.09);
        \draw[red!75!black, line width=1.0pt, line cap=round] (-0.09, 0.09) -- (0.09,-0.09);
    }},
    pics/imgcell/.style 2 args={code={
        \draw[gray!55] (0,0) rectangle (1.6, 1.6);
        \draw[gray!55] (0.8, 0) -- (0.8, 1.6);
        \draw[gray!55] (0, 0.8) -- (1.6, 0.8);
    }},
}

\node[font=\footnotesize, align=center, inner sep=0pt] at (1.3, 6.85) {Image 1\\[-2pt]tokens};

\begin{scope}[shift={(0.5, 4.6)}]
    \draw[gray!55] (0,0) rectangle (1.6, 1.6);
    \draw[gray!55] (0.8, 0) -- (0.8, 1.6);
    \draw[gray!55] (0, 0.8) -- (1.6, 0.8);
    \path (0.4, 1.2) pic{circletok};
    \path (1.2, 1.2) pic{circletok};
    \path (0.4, 0.4) pic{circletok};
    \path (1.2, 0.4) pic{circletok};
\end{scope}

\begin{scope}[shift={(0.5, 1.5)}]
    \draw[gray!55] (0,0) rectangle (1.6, 1.6);
    \draw[gray!55] (0.8, 0) -- (0.8, 1.6);
    \draw[gray!55] (0, 0.8) -- (1.6, 0.8);
    \foreach \x/\y in {0.4/1.2, 1.2/1.2, 0.4/0.4, 1.2/0.4}
        \path (\x, \y) pic{redtriangletok};
\end{scope}

\node[font=\footnotesize, align=center, inner sep=0pt] at (1.3, 3.75) {Image 2\\[-2pt]tokens};

\begin{scope}[shift={(3.5, 2.5)}, scale=1.0]

    \begin{scope}
        \clip (-1.0, -0.8) rectangle (4.0, 5.0);
        \input{figures_arxiv/method_illustration/gmm_contours_s2}
    \end{scope}

    \path (0.4, 0.3)  pic{redtriangletok};
    \path (-0.3, 2.6) pic{redtriangletok};
    \path (2.4, 3.4)  pic{redtriangletok};
    \path (2.3, 2.0)  pic{redtriangletok};

    \path (0, 0)        pic{circletok};
    \path (-0.3, 2.824) pic{circletok};
    \path (3, 2.7)        pic{circletok};
    \path (3, 0.5)      pic{circletok};

    \draw[push_arrow] (0,    0)   -- (-0.56, -0.42);
    \draw[push_arrow] (-0.3, 2.824) -- (-0.3, 3.524);
    \draw[push_arrow] (3,  2.7)   -- ( 3.7,  2.7);

\end{scope}

\end{tikzpicture}}}\hspace{6pt}%
            \adjustbox{max width=0.49\linewidth, max totalheight=0.333\linewidth}{\resizebox{0.49\linewidth}{!}{\begin{tikzpicture}[
    scale=0.3,
    >=Stealth,
    every node/.append style={font=\scriptsize},
    every pic quotes/.append style={font=\scriptsize},
    sec3_vec/.style      ={-{Stealth[length=4pt]}, thick, black},
    sec3_redvec/.style   ={-{Stealth[length=4pt]}, thick, red!80!black},
    sec3_pushvec/.style  ={-{Stealth[length=4pt]}, thick, blue!65!black},
    sec3_rotarrow/.style ={-{Stealth[length=3.5pt]}, thick, red!65!black},
]

\coordinate (O)    at (0, 0);
\coordinate (H)    at (0, 5);
\coordinate (Href) at (2.11, 4.53);
\coordinate (Hp)   at (-3.75, 5);

\draw[gray!45, dashed, thin]
    ({5*cos(-22)}, {5*sin(-22)}) arc[start angle=-22, end angle=202, radius=5];

\draw[sec3_vec]     (O) -- (H);
\draw[sec3_vec]     (O) -- (Href);
\draw[sec3_redvec]  (O) -- (Hp);
\draw[sec3_pushvec] (H) -- (Hp);

\node[above right=-1pt]                 at (H)    {$\hat{\mathbf{h}}_{ij}$};
\node[right=2pt]                        at (Href) {$\mubar(\hat{\mathbf{h}}_{ij})$};
\node[above left=-4pt, red!80!black]    at (Hp)   {$\mathbf{h}'_{ij}$};
\node[above=4pt, blue!65!black]         at ($(H)!0.58!(Hp)$) {$-\lambda \nabla p (\hat{\mathbf{h}}_{ij})$};

\pic [draw, thin, angle radius=0.5cm,
      "$\theta$", angle eccentricity=1.6]
     {angle = Href--O--H};

\pic [draw, thin, red!70!black, angle radius=0.5cm,
      "$\delta$" {red!70!black}, angle eccentricity=1.6]
     {angle = H--O--Hp};

\fill (O) circle (1.5pt);

\end{tikzpicture}}}
            \caption{Token repulsion}
            \label{fig:repel_repulsion}
        \end{subfigure}

        \vspace{4pt}
        \begin{subfigure}[b]{\linewidth}
            \centering
            \adjustbox{max width=0.49\linewidth, max totalheight=0.333\linewidth}{\resizebox{0.49\linewidth}{!}{\begin{tikzpicture}[
    scale=0.65, 
    >=Stealth,
    vec/.style      ={-{Stealth[length=8pt]}, line width=1.1pt, black},
    greenvec/.style ={-{Stealth[length=8pt]}, line width=1.1pt, dashed, green!40!black},
]

\begin{scope}
    \clip (-1, -1) rectangle (3.5, 3.1);
    \input{figures_arxiv/method_illustration/gmm_contours_score}
\end{scope}

\draw[vec] (-0.5,-0.5) -- (-0.11,0.95);

\draw[greenvec] (-0.5,-0.5) -- (0.7,0.8);

\node[anchor=east, font=\footnotesize] at (-0.4, 0.3) {$\nabla \log p^r_t(x_t)$};
\node[anchor=west, font=\footnotesize, green!40!black] at (0.28, -0.02) {$\nabla \log p_t(x_t)$};

\fill (-0.5,-0.5) circle (2pt);

\end{tikzpicture}}}\hspace{6pt}%
            \adjustbox{max width=0.49\linewidth, max totalheight=0.333\linewidth}{\resizebox{0.49\linewidth}{!}{\begin{tikzpicture}[
    scale=0.29,
    yscale=1.38,
    >=Stealth,
    every node/.append style={font=\footnotesize},
    vec/.style        ={-{Stealth[length=9pt]}, line width=1.2pt, black},
    grayvec/.style    ={-{Stealth[length=9pt]}, line width=1.2pt, dashed, black!60},
    greenvec/.style   ={-{Stealth[length=9pt]}, line width=1.2pt, dashed, green!40!black},
    bluevec/.style  ={-{Stealth[length=9pt]}, line width=1.2pt, dashed, blue!60!black},
    redvec/.style     ={-{Stealth[length=9pt]}, line width=1.2pt, red!80!black},
    dashline/.style   ={dashed, gray!70, thick},
    rightangle/.style ={gray!70, thin},
]

\fill[orange!7] (5,-2.2) rectangle (7.5,4);
\path (5,-4);

\draw[greenvec] (-4,0) -- (5,0);
\node[below=3pt, green!40!black] at (1, 0) {$\nabla \log p_t(x_t)$};
\draw[rightangle] (4.6, 0) -- (4.6, 0.4) -- (5, 0.4);

\draw[bluevec] (5,0) -- (3,3);
 \node[right=50pt, blue!50!black] at (-2.2, 2.0) {$ \propto \Delta^r$};

\draw[dashline] (3,3) -- (5,3);

\draw[rightangle] (4.6, 3) -- (4.6, 3.4) -- (5, 3.4);

\draw[vec]    (-4,0) -- (3,3)
    node[midway, above left=-2pt] {$\nabla \log p^r_t(x_t)$};

\draw[redvec] (-4,0) -- (5,3);

\fill (-4,0) circle (1.5pt);
\fill (5,0)  circle (1.5pt);

\end{tikzpicture}}}
            \caption{Token density preservation}
            \label{fig:repel_preservation}
        \end{subfigure}
    \end{minipage}\hfill
    \begin{minipage}[b]{0.32\textwidth}
        \begin{subfigure}[b]{\linewidth}
            \centering
            \adjustbox{max width=\linewidth}{\begin{tikzpicture}[
    >=Stealth,
    every node/.append style={font=\footnotesize},
    tok/.style={draw=#1!70!black, fill=#1!25, minimum size=0.42cm, inner sep=0pt, line width=0.4pt},
    stage/.style={font=\footnotesize\bfseries},
    op/.style={draw=#1!70!black, fill=#1!8, rounded corners=2pt, thick, minimum width=3.2cm, minimum height=0.55cm, inner xsep=3pt, align=center},
    flow/.style={->, thick, gray!70!black},
]

\newcommand{\tokenrow}[2]{%
    \foreach \b in {0,1} {
        \foreach \k in {0,...,3} {
            \node[tok=#2] at ({0.21+\k*0.42+\b*1.98}, #1) {};
        }
    }
    \foreach \d in {0,1,2} {
        \fill[black] ({3.86+\d*0.13}, #1) circle (0.03);
    }
}

\newcommand{\layerarrow}[3]{%
    \draw[fill=gray!40, draw=gray!60!black, line width=0.4pt]
        (1.75, #1) -- (1.91, #1) -- (1.91, {#2+0.2}) -- (2.0, {#2+0.2})
        -- (1.83, #2) -- (1.66, {#2+0.2}) -- (1.75, {#2+0.2}) -- cycle;
    \node[right] at (2.05, {(#1+#2)/2}) {#3};
}

\node[stage] at (1.83, 1.12) {Hidden states};
\tokenrow{0}{blue}
\node[align=center, inner sep=0pt] at (0.84, 0.52) {Image 1\\[-2pt]tokens};
\node[align=center, inner sep=0pt] at (2.82, 0.52) {Image 2\\[-2pt]tokens};
\layerarrow{-0.31}{-0.81}{Attention}

\newcommand{\plusop}[1]{%
    \draw[flow, line width=1.2pt, -{Stealth[length=5pt, width=5.5pt]}] (1.83, {#1+0.2}) -- (1.83, {#1-0.2});
}

\node[stage] at (1.83, -1.06) {Attention output};
\tokenrow{-1.46}{orange}
\plusop{-1.91}

\node[op=red] at (1.83, -2.43) {Token repulsion};
\layerarrow{-2.81}{-3.31}{MLP + Residual}

\node[stage] at (1.83, -3.56) {Layer output};
\tokenrow{-3.96}{green}
\plusop{-4.41}

\node[op=blue] at (1.83, -4.93) {Token density preservation};

\end{tikzpicture}}
            \caption{Overall pipeline}
            \label{fig:repel_pipeline}
        \end{subfigure}
    \end{minipage}
    
    \caption{Illustration of our TokenRepel algorithm. 
    \subref{fig:repel_repulsion}~Left: The tokens of Image 1 are repelled by taking a negative gradient step induced by the distribution formed from the tokens of the other images.
    \subref{fig:repel_repulsion}~Right: Update formulation for $\mathbf{h}_{ij}$, 
    where $\bar{\mu}(\hat{\mathbf{h}}_{ij})$ denotes the weighted mean of the other image tokens with respect to their cosine similarity to $\mathbf{h}_{ij}$.
    \subref{fig:repel_preservation}~Left: Contours representing the learned data distribution of the generative model, a green dashed arrow representing the original score, and a black solid arrow representing the score after applying token repulsion. 
    \subref{fig:repel_preservation}~Right: The refinement process applied to $\nabla \log p^r_t(x_t)$ to project it into the higher-density half-space (red solid arrow).    
    \subref{fig:repel_pipeline}~Locations of these two operations inside a transformer layer.
    }
    \label{fig:repel}
   
\end{figure}

\section{Enhancing Image Diversity}
\label{sec:method_diversity}
In this section, we present our algorithm for enhancing image diversity, as depicted in \cref{fig:repel_repulsion}.
Given a set of image token features, our goal is to have similar features repel each other.
This can be visualized as an electron cloud (feature set) in which electrons (features) repel one another when too close but are otherwise unaffected.
Like electrons around a nucleus, the features cannot move arbitrarily; they remain on spherical shells.
This reflects the notion that the angle between two feature vectors captures their semantic similarity more faithfully than Euclidean distance \cite{dreammover, negtome, corres_diffusion}; 
thus, repulsion should change their angle rather than their magnitudes.
The PyTorch-style pseudocode is provided in \cref{alg:token_repulsion}.
\begin{algorithm}[t]
    \caption{PyTorch-style pseudocode of token repulsion.}
    \label{alg:token_repulsion}
    \definecolor{codegreen}{rgb}{0.1,0.6,0.1}
    \definecolor{codepurple}{rgb}{0.85,0.18,0.50}
    \lstset{
        backgroundcolor=\color{white},
        basicstyle=\fontsize{7.5pt}{8pt}\ttfamily\selectfont,
        columns=fullflexible,
        breaklines=true,
        commentstyle=\color{codegreen},
        keywordstyle=\color{codepurple},
        showstringspaces=false,
    }
    \begin{lstlisting}[language=python]
def TokenRepel(h, delta, theta_bar):
    # h: [B, N, D] image tokens after an attention layer
    # delta: token rotation angle
    # theta_bar: maximum angle for receiving repulsion
    B, N, D = h.shape
    h = h.reshape(B * N, D)
    h_hat = F.normalize(h, dim=-1)

    # 1) Most similar token from the other images
    image_id = torch.arange(B).repeat_interleave(N)
    cosine = h_hat @ h_hat.T
    cosine[image_id[:, None] == image_id[None, :]] = -inf
    max_cos, idx = cosine.max(dim=-1)
    max_cos = max_cos[:, None]
    h_ref = h_hat[idx]

    # 2) Unit tangent direction pointing away from h_ref
    #    (a random tangent direction if h_hat and h_ref coincide)
    t = F.normalize(max_cos * h_hat - h_ref, dim=-1)

    # 3) Rotate on the unit sphere without passing 90 degrees
    h_new = h_hat + math.tan(delta) * t
    overshoot = (h_new * h_ref).sum(-1, keepdim=True) < 0
    h_new = torch.where(overshoot, h_hat - max_cos * h_ref, h_new)
    h_new = F.normalize(h_new, dim=-1) * h.norm(dim=-1, keepdim=True)

    # 4) Repel only tokens above the similarity threshold
    h_new = torch.where(max_cos > math.cos(theta_bar), h_new, h)
    return h_new.reshape(B, N, D)
    \end{lstlisting}
    \vspace{-0.55em}
\end{algorithm}

\paragraph{Modeling cross-image token interactions.}
We model the normalized token features from all other images as a probability density on the unit sphere using a von Mises--Fisher (vMF) mixture model.
Let $\mathbf{h}_{ij} \in \mathbb{R}^{D}$ denote the hidden state of the $j$\textsuperscript{th} token in the $i$\textsuperscript{th} image after a given attention layer, and let $\hhij = \mathbf{h}_{ij}/\|\mathbf{h}_{ij}\|$ be its unit-norm counterpart.
Here, $i \in [1,B]$ and $j \in [1,N]$ for a batch of $B$ images with $N$ tokens per image.
We model the density of the normalized hidden state features as an isotropic equal-weight von Mises--Fisher (vMF) mixture with probability density function
\begin{align}
    p(\hhij) = \frac{C_D(\kappa)}{(B{-}1)N} \sum_{k \neq i,\, \ell} \exp\bigl( \kappa\, \hhij^\mathsf{T} \hhkl \bigr),
\end{align}
where $C_D(\kappa)$ is the vMF normalizing constant in $D$ dimensions, $\kappa$ is the concentration parameter, and the mixture means $\hhkl$ are the $(B-1)N$ features from images other than $I_i$.
This corresponds to a kernel density estimate of the distribution.
The Riemannian gradient of the probability density with respect to $\hhij$ is
\begin{align}
    \nabla p(\hhij) 
    =  p(\hhij) \nabla \log p(\hhij) 
    = \kappa p(\hhij) T(\mubar(\hhij), \hhij)
    \label{eq:vmf_gradient}
\end{align}
where
$T(\mathbf{a},\mathbf{b})= \mathbf{a} - (\mathbf{a}^\mathsf{T} \mathbf{b})\, \mathbf{b}$
is the tangent space projection operator and
\begin{align}
    \boldsymbol{\bar{\mu}}(\hhij) &= \sum_{k \neq i, \ell} \frac{\exp{(\kappa \hhij^\mathsf{T} \hhkl)}}{\sum_{m\neq i,n} \exp{(\kappa \hhij^\mathsf{T} \hat{\mathbf{h}}_{mn})}} \hhkl
    \label{eq:vmf_softmax_mean}
\end{align}
is the softmax-weighted mean of the other features.
Geometrically, $\nabla p(\hhij)$ is proportional to the projection of $\boldsymbol{\bar{\mu}}(\hhij) - \hhij$ 
onto the tangent space at $\hhij$.

\paragraph{Token repulsion.}
In the high-dimensional token spaces of generative models (e.g., $D=3072$ for FLUX.1-dev), the vMF density $p(\hhij)$ is sharply concentrated and decays rapidly as the angular separation $\theta$ increases.
Retaining interactions over a useful angular range therefore requires a large concentration $\kappa$.
For example, placing the transition near $\theta=45^\circ$ in $D=3072$ requires $\kappa$ on the order of $50{,}000$.
At this concentration, the density is effectively step-like, falling from a large value to nearly zero, while the softmax-weighted mean $\mubar(\hhij)$ approaches argmax.
We therefore closely approximate the update by a uniform repulsion from the most similar cross-image token when their angular separation is below a threshold $\bar{\theta}$, and no update otherwise.

To do so, we leverage two geometrically meaningful parameters: the token rotation angle $\delta$ and the angular threshold $\bar{\theta}$, 
instead of the step size $\lambda$ and concentration $\kappa$.
The token update is formulated as:
\begin{align}
    \hij' &= \hhij - \lambda \nabla p(\hhij)
    \approx \hhij - \tan(\delta)
    \mathbf{1}[\hhij^\mathsf{T}\hhklmax > \cos(\bar{\theta})]
    T(\hhklmax,\hhij) / \|T(\hhklmax,\hhij)\|,
\end{align}
where $k^\star, \ell^\star = \arg\max_{k\neq i,\,\ell}\, \hhij^\mathsf{T} \hhkl$.
If $\theta=0$, the two tokens coincide, the norm is zero and we replace this term with a random unit direction in the tangent space at $\hhij$.
We also cap the rotation so that $\bar{\theta}+\delta\leq90^\circ$, preventing the update from overshooting orthogonality.

\section{Preserving Image Fidelity}
\label{sec:method_fidelity}

An unconstrained token repulsion can improve diversity but may move predictions away from the model's learned distribution, 
reducing image quality and prompt adherence.
To preserve fidelity, we introduce a token density preservation mechanism that prevents repulsion-induced updates from pointing away from high-density regions of latent space.
The pretrained flow-matching model implicitly encodes a direction toward these regions \cite{density_geodesics,score_sde,ncsn}, 
which we infer from its model predictions and use to restrict the token movement.
The PyTorch-style pseudocode is provided in \cref{alg:density_preservation}.
\begin{algorithm}[t]
    \caption{PyTorch-style pseudocode of token density preservation.}
    \label{alg:density_preservation}
    \definecolor{codegreen}{rgb}{0.1,0.6,0.1}
    \definecolor{codepurple}{rgb}{0.85,0.18,0.50}
    \lstset{
        backgroundcolor=\color{white},
        basicstyle=\fontsize{7.5pt}{8pt}\ttfamily\selectfont,
        columns=fullflexible,
        keepspaces=true,
        breaklines=true,
        commentstyle=\color{codegreen},
        keywordstyle=\color{codepurple},
        showstringspaces=false,
    }
    \begin{lstlisting}[language=python]
def Preserve(u_r, u_pos, u_neg):
    # u_r, u_pos, u_neg: [B, N, D] outputs with TokenRepel,
    # with prompt c, and with negative prompt c_neg
    d_r = u_r - u_pos                                    # repulsion-induced change
    d_h = F.normalize(u_pos - u_neg, dim=-1)             # high-density direction
    proj = (d_r * d_h).sum(-1, keepdim=True)
    d_r = torch.where(proj < 0, d_r - proj * d_h, d_r)   # half-space projection
    return u_pos + d_r
    \end{lstlisting}
    \vspace{-1.3em}
\end{algorithm}

\paragraph{Token density preservation.}
Ideally, we would compare the score of the original distribution,
$\nabla_{\xbf_t}\log p_t(\xbf_t \mid c)$,
with the score after token repulsion,
$\nabla_{\xbf_t}\log p_t^r(\xbf_t \mid c)$.
Here, $\xbf_t$ is the latent at time $t$ and $c$ is the text condition.
We seek to prevent the repulsion-induced change $\sbf^r-\sbf$ from opposing a direction of increasing density.
However, a flow-matching model predicts a velocity $\vbf_t(\xbf_t,c)$ rather than either score directly.
For the model's probability path, the score can be expressed as a time-dependent linear combination of the velocity and the latent,
\begin{align}
    \nabla_{\xbf_t}\log p_t(\xbf_t \mid c) \propto \vbf_t(\xbf_t,c) + a_t\xbf_t,
\end{align}
where $a_t$ depends only on time.
At the same $\xbf_t$, the latent term cancels when we subtract two predictions, making velocity differences proportional to score differences.
We therefore define
\begin{align}
\Delta^{r} &\triangleq \vbf_t^r(\xbf_t, c) - \vbf_t(\xbf_t, c) \varpropto \nabla_{\xbf_t}\log p_t^r(\xbf_t \mid c) - \nabla_{\xbf_t}\log p_t(\xbf_t \mid c),\\
\Delta^{h} &\triangleq \vbf_t(\xbf_t, c) - \vbf_t(\xbf_t, c_\text{neg}) \varpropto \nabla_{\xbf_t}\log p_t(\xbf_t \mid c) - \nabla_{\xbf_t}\log p_t(\xbf_t \mid c_\text{neg}),
\end{align}
where $\vbf_t^r$ is the velocity prediction with TokenRepel and $\nabla_{\xbf_t}\log p_t(\xbf_t \mid c_\text{neg})$ is the score under a negative prompt $c_\text{neg}$.
Thus, $\Delta^r$ captures the repulsion-induced change, while $\Delta^h$ provides a direction toward higher density under the desired condition relative to the negative prompt.
At the token level, $\Delta^r$ and $\Delta^h$ become more negatively aligned as the repulsion angle increases, as shown in \cref{fig:token_density_alignment_sigma}, 
motivating the following correction mechanism.

For the $j$\textsuperscript{th} token of the $i$\textsuperscript{th} image, we constrain the repulsion-induced change to the half-space
$\mathcal{R}_{ij} = \{\mathbf{d} : \langle \mathbf{d}, \Delta^h_{ij} \rangle \geq 0\}$.
Its boundary is orthogonal to $\Delta^h_{ij}$, as illustrated in \cref{fig:repel_preservation}.
As derived in \cref{app:token_density_constraint}, if $\Delta^r_{ij}$ points outside this half-space, we remove its component opposing $\Delta^h_{ij}$; otherwise, we leave it unchanged:
\begin{align}
    \bar{\Delta}^r_{ij} =
    \begin{cases}
    \displaystyle 
    \Delta^r_{ij} - \langle \Delta^r_{ij}, \widehat{\Delta}^h_{ij} \rangle\, \widehat{\Delta}^h_{ij}
    & \text{if } \langle \Delta^r_{ij}, \Delta^h_{ij} \rangle < 0, \\[6pt]
    \Delta^r_{ij} & \text{otherwise,}
    \end{cases}
    \label{eq:repel_plus_correct}
\end{align}
where $\widehat{\Delta}^h_{ij} = \Delta^h_{ij} / \|\Delta^h_{ij}\|$.
This is the Euclidean projection of $\Delta^r_{ij}$ onto $\mathcal{R}_{ij}$.
Denoting this constraint by $\phi$, the updated velocity is
\begin{align}
   \tilde{\vbf}^r = \phi (\vbf^r(\xbf, c), \vbf(\xbf, c), \vbf(\xbf, c_{\text{neg}})) = \vbf(\xbf, c) + \bar{\Delta}^r,
\end{align}
We use $\tilde{\vbf}^r$ to obtain the next latent $\xbf_{t-1}$.
Thus, the velocity prediction after applying token repulsion is constrained to the high-density half-space.
For $\epsilon$-prediction models such as SDXL, the same constraint formulation applies.
It better steers the token updates to the high-density region while enabling sufficient diversity. 

\paragraph{Layerwise density preservation.}
\label{sec:method_layers}
Applying token density preservation only at the final block output (velocities) preserves image quality, 
but repulsion-induced errors can accumulate excessively across layers.
We therefore correct the intermediate hidden states as repulsion proceeds, 
using smaller, more frequent adjustments to guide them toward high-density regions.

Given a model with $M$ transformer layers, the latent embedding $\xbf$ is first projected into a higher-dimensional hidden space, 
passed through the $M$ layers, and finally projected back to the dimension of $\xbf$ to predict the velocity.
The process can be written as
\begin{align}
    \ubf_1 = f_{\uparrow}(\xbf), \qquad \ubf_{m} = \ubf_{m-1} + g_m{(\ubf_{m-1})}, \qquad \vbf = f_{\downarrow}(\ubf_M)   
\end{align}
where $f_{\uparrow}$ and $f_{\downarrow}$ are the up-projection and down-projection transformations, and $g_m$ represents the $m$th transformer layer.
Each transformer layer produces the next hidden state through a residual connection.
Thus, each successive hidden state can be viewed as a closer approximation of $\ubf_M$, which is mapped to the velocity $\vbf$.

Therefore, for each layer, we maintain three parallel streams:
one using the provided prompt,
another using the negative prompt,
and a third using the provided prompt with TokenRepel enabled.
We denote their hidden states after the $m$th layer by $\ubf^+_m$, $\ubf^-_m$, and $\ubf^r_m$, respectively.
Using the constraint operator $\phi$ defined above, we replace $\ubf^r_m$ with
\begin{align}
    \tilde{\ubf}^r_m = \phi (\ubf^r_m, \ubf^+_m, \ubf^-_m).
\end{align}
This layerwise preservation better steers token updates towards high-density regions while preserving sufficient diversity.

\section{Experiments}
We evaluate TokenRepel on three representative T2I models: 
FLUX.1-dev~\cite{flux} (flow matching), FLUX.2-Klein-9B~\cite{flux2klein} (few-step distilled flow matching), and SDXL~\cite{sdxl} (diffusion).
We report all results on the GenEval dataset~\cite{geneval}.

\paragraph{Implementation details.}
For FLUX.1-dev, FLUX.2-Klein-9B, and SDXL, we use 25, 4, and 50 inference steps,
and apply TokenRepel during the first 3, 1, and 5 steps, respectively;
repulsion is applied in all layers, and density preservation is applied only in the early layers (first 30, 20, and downsample \& middle layers).
We apply TokenRepel only to the conditional branch on SDXL.
We generate 4 images per prompt at a resolution of 512 and use the same initial noise across methods for each prompt.
Further implementation details and parameter settings are provided in \cref{app:implementation_details}.

\paragraph{Evaluation metrics.}
For diversity evaluation, we use three metrics:
(1) DreamSim~\cite{dreamsim}, trained on human-labeled data to measure visual differences, where we use the ensemble version (CLIP, DINO, and OpenCLIP; all ViT-B/16) and report the mean pairwise distance within each prompt batch;
(2) Vendi Score~\cite{vendi}, which computes the Shannon entropy of the eigenvalues of the cosine similarity matrix of InceptionV3 features; and
(3) LPIPS~\cite{lpips} (VGG backbone), which measures pairwise perceptual distances within each batch.
For image quality, we use multiple reward models capturing different aspects:
(1) HPSv3~\cite{hpsv3} (Qwen-7B-based), which evaluates overall quality and prompt alignment;
(2) ImageReward~\cite{imagereward} (BLIP-based), which assesses overall quality and alignment;
(3) CLIP~\cite{clip} (ViT-H/14), which focuses on prompt adherence;
(4) PickScore~\cite{pickapic}, a CLIP-based human preference model; and
(5) GenEval~\cite{geneval}, which measures compositionality and constraint satisfaction using a detection model.

\subsection{Main Evaluation}
\label{exp:compare}

We compare TokenRepel with training-free methods for enhancing image diversity, including
CADS~\cite{cads}, SPARKE~\cite{sparke}, Particle Guidance (PG)~\cite{particle_g}, NegToMe~\cite{negtome}, Contextual Repulsion~\cite{contextual_repulsion}, Group Inference~\cite{group_inference}, Shielded Diffusion~\cite{shielded_diffusion}, and Contrastive Noise Optimization (CNO)~\cite{cno}.
Each method has a hyperparameter that controls the diversity--quality trade-off.
We sweep this parameter on GenEval for FLUX.1-dev, FLUX.2-Klein-9B, and SDXL.

\paragraph{Comparison results.}
\Cref{fig:geneval_swap_comparison} shows the trade-off curves using DreamSim and HPSv3, our most comprehensive diversity and image quality metrics, respectively.
Full evaluations of every sweep point are reported in \cref{app:geneval_sweep}.
We sweep four hyperparameter values for each method, with the HPSv3 drop roughly capped at 1.0.
The swept values and fixed settings of each method are listed in \cref{app:parameter_settings}.
TokenRepel gives the best trade-off on FLUX.2-Klein-9B and the second-best trade-off on FLUX.1-dev and SDXL.
Its performance is also the most stable across the three models.
For each method, we select the setting that maximizes DreamSim
while keeping HPSv3 within a model-specific range,
which reflects the variation of baseline HPSv3 across random seeds.
\Cref{tab:comparison} reports the full evaluation metrics at the selected settings and their latency.
TokenRepel achieves the highest DreamSim on both FLUX models and the second-highest on SDXL while maintaining image quality metrics close to the baseline.
The implementation details of all methods are provided in \cref{app:implementation_details}.
Qualitative examples at the selected settings are presented in \cref{fig:qual_geneval_flux,fig:qual_geneval_flux2k,fig:qual_geneval_sdxl}.

\pgfdeclareplotmark{filled pentastar}{%
  \pgfmathsetlengthmacro{\pentastarinner}{0.382*\pgfplotmarksize}%
  \pgfpathmoveto{\pgfqpointpolar{90}{\pgfplotmarksize}}%
  \pgfpathlineto{\pgfqpointpolar{54}{\pentastarinner}}%
  \pgfpathlineto{\pgfqpointpolar{18}{\pgfplotmarksize}}%
  \pgfpathlineto{\pgfqpointpolar{-18}{\pentastarinner}}%
  \pgfpathlineto{\pgfqpointpolar{-54}{\pgfplotmarksize}}%
  \pgfpathlineto{\pgfqpointpolar{-90}{\pentastarinner}}%
  \pgfpathlineto{\pgfqpointpolar{-126}{\pgfplotmarksize}}%
  \pgfpathlineto{\pgfqpointpolar{-162}{\pentastarinner}}%
  \pgfpathlineto{\pgfqpointpolar{162}{\pgfplotmarksize}}%
  \pgfpathlineto{\pgfqpointpolar{126}{\pentastarinner}}%
  \pgfpathclose\pgfusepathqfillstroke%
}
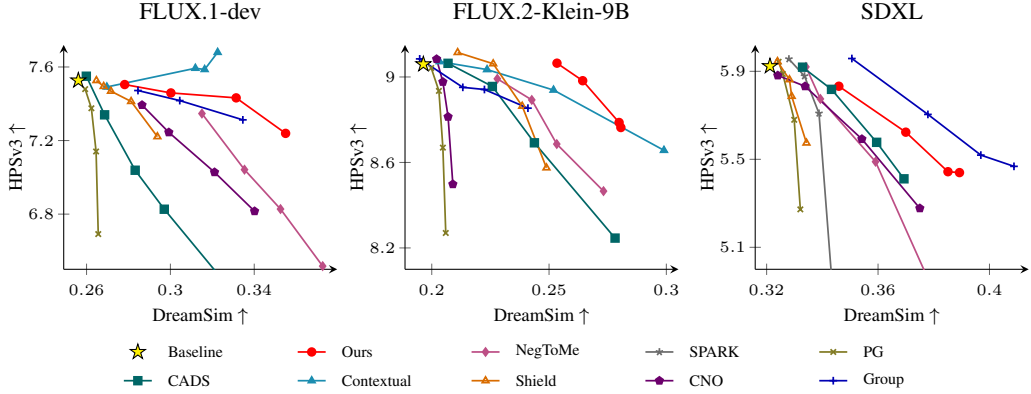
\begin{figure}[t]
\centering
\begin{subfigure}[t]{0.34\textwidth}
\centering
\begin{tikzpicture}[font=\fontfamily{phv}\selectfont]
\begin{axis}[
  width=5.20cm, height=4.55cm,
  title={FLUX.1-dev}, title style={font=\small},
  label style={font=\scriptsize}, tick label style={font=\tiny},
  axis lines=left, scaled ticks=false,
  xticklabel style={/pgf/number format/fixed, /pgf/number format/precision=2},
  yticklabel style={/pgf/number format/fixed, /pgf/number format/precision=1, xshift=2pt},
  xmin=0.2491, xmax=0.3796,
  xtick={0.26,0.3,0.34},
  ymin=6.50, ymax=7.72,
  ytick={6.8,7.2,7.6},
  tick style={black!60},
  extra description/.code={%
    \node[font=\scriptsize, rotate=90, anchor=center] at (axis description cs:-0.17,0.5) {HPSv3 $\uparrow$};%
    \node[font=\scriptsize, anchor=center] at (axis description cs:0.5,-0.21) {DreamSim $\uparrow$};%
  },
]
\addplot[color=red, mark=*, mark size=1.45pt, line width=0.65pt] coordinates {(0.27819,7.50453) (0.30024,7.45894) (0.33138,7.43207) (0.35497,7.23905)};
\addplot[color=magenta!75!black, mark=diamond*, mark size=1.45pt, line width=0.65pt] coordinates {(0.31505,7.34673) (0.33536,7.04173) (0.35254,6.82749) (0.37265,6.51729)};
\addplot[color=olive!80!black, mark=x, mark size=1.45pt, line width=0.65pt] coordinates {(0.25934,7.47964) (0.26231,7.37559) (0.26455,7.14065) (0.26558,6.69147)};
\addplot[color=teal!80!black, mark=square*, mark size=1.45pt, line width=0.65pt] coordinates {(0.25989,7.54981) (0.26862,7.33987) (0.28320,7.03863) (0.29708,6.82663) (0.32926,6.38290)};
\addplot[color=cyan!65!black, mark=triangle*, mark size=1.45pt, line width=0.65pt] coordinates {(0.26968,7.49127) (0.31192,7.59307) (0.31634,7.58639) (0.32254,7.67952)};
\addplot[color=orange!85!black, mark=triangle, mark size=1.45pt, line width=0.65pt] coordinates {(0.26470,7.52626) (0.26817,7.49448) (0.27165,7.46833) (0.28126,7.41266) (0.29380,7.22120)};
\addplot[color=violet!85!black, mark=pentagon*, mark size=1.45pt, line width=0.65pt] coordinates {(0.28647,7.39307) (0.29925,7.24457) (0.32106,7.02777) (0.34017,6.81579)};
\addplot[color=blue!70!black, mark=+, mark size=1.45pt, line width=0.65pt] coordinates {(0.28453,7.47201) (0.30447,7.41728) (0.33457,7.31255)};
\addplot[only marks, draw=black, fill=yellow, mark=filled pentastar, mark size=3pt] coordinates {(0.25608,7.52661)};
\end{axis}
\end{tikzpicture}
\end{subfigure}
\hspace{-1em}
\begin{subfigure}[t]{0.34\textwidth}
\centering
\begin{tikzpicture}[font=\fontfamily{phv}\selectfont]
\begin{axis}[
  width=5.20cm, height=4.55cm,
  title={FLUX.2-Klein-9B}, title style={font=\small},
  label style={font=\scriptsize}, tick label style={font=\tiny},
  axis lines=left, scaled ticks=false,
  xticklabel style={/pgf/number format/fixed, /pgf/number format/precision=2},
  yticklabel style={/pgf/number format/fixed, /pgf/number format/precision=1, xshift=2pt},
  xmin=0.1886, xmax=0.3052,
  ymin=8.10, ymax=9.15,
  ytick={8.2,8.6,9.0},
  tick style={black!60},
  extra description/.code={%
    \node[font=\scriptsize, rotate=90, anchor=center] at (axis description cs:-0.17,0.5) {HPSv3 $\uparrow$};%
    \node[font=\scriptsize, anchor=center] at (axis description cs:0.5,-0.21) {DreamSim $\uparrow$};%
  },
]
\addplot[color=red, mark=*, mark size=1.45pt, line width=0.65pt] coordinates {(0.25338,9.06456) (0.26434,8.98247) (0.27986,8.78705) (0.28061,8.76340)};
\addplot[color=magenta!75!black, mark=diamond*, mark size=1.45pt, line width=0.65pt] coordinates {(0.22795,8.99152) (0.24264,8.89261) (0.25329,8.68605) (0.27322,8.46582)};
\addplot[color=olive!80!black, mark=x, mark size=1.45pt, line width=0.65pt] coordinates {(0.19977,9.04210) (0.20305,8.93475) (0.20474,8.66935) (0.20597,8.26991)};
\addplot[color=teal!80!black, mark=square*, mark size=1.45pt, line width=0.65pt] coordinates {(0.20694,9.06388) (0.22583,8.95554) (0.24376,8.69189) (0.27814,8.24599)};
\addplot[color=cyan!65!black, mark=triangle*, mark size=1.45pt, line width=0.65pt] coordinates {(0.20337,9.07186) (0.22358,9.03454) (0.25191,8.93872) (0.29899,8.65669)};
\addplot[color=orange!85!black, mark=triangle, mark size=1.45pt, line width=0.65pt] coordinates {(0.21104,9.11454) (0.22619,9.06184) (0.23857,8.86339) (0.24890,8.57485)};
\addplot[color=violet!85!black, mark=pentagon*, mark size=1.45pt, line width=0.65pt] coordinates {(0.20206,9.08363) (0.20474,8.97694) (0.20696,8.81355) (0.20899,8.49813)};
\addplot[color=blue!70!black, mark=+, mark size=1.45pt, line width=0.65pt] coordinates {(0.19486,9.08458) (0.21324,8.95149) (0.22246,8.94141) (0.24100,8.85403)};
\addplot[only marks, draw=black, fill=yellow, mark=filled pentastar, mark size=3pt] coordinates {(0.19642,9.05947)};
\end{axis}
\end{tikzpicture}
\end{subfigure}
\hspace{-1em}
\begin{subfigure}[t]{0.34\textwidth}
\centering
\begin{tikzpicture}[font=\fontfamily{phv}\selectfont]
\begin{axis}[
  width=5.20cm, height=4.55cm,
  title={SDXL}, title style={font=\small},
  label style={font=\scriptsize}, tick label style={font=\tiny},
  axis lines=left, scaled ticks=false,
  xticklabel style={/pgf/number format/fixed, /pgf/number format/precision=2},
  yticklabel style={/pgf/number format/fixed, /pgf/number format/precision=1, xshift=2pt},
  xmin=0.3159, xmax=0.4141,
  xtick={0.32,0.36,0.4},
  ymin=5.00, ymax=6.02,
  ytick={5.1,5.5,5.9},
  tick style={black!60},
  extra description/.code={%
    \node[font=\scriptsize, rotate=90, anchor=center] at (axis description cs:-0.17,0.5) {HPSv3 $\uparrow$};%
    \node[font=\scriptsize, anchor=center] at (axis description cs:0.5,-0.21) {DreamSim $\uparrow$};%
  },
]
\addplot[color=red, mark=*, mark size=1.45pt, line width=0.65pt] coordinates {(0.34593,5.83220) (0.36996,5.62335) (0.38510,5.44366) (0.38924,5.43960)};
\addplot[color=magenta!75!black, mark=diamond*, mark size=1.45pt, line width=0.65pt] coordinates {(0.33410,5.92010) (0.33930,5.77450) (0.35910,5.48900) (0.38300,4.81330)};
\addplot[color=gray!85!black, mark=star, mark size=1.45pt, line width=0.65pt] coordinates {(0.32800,5.95580) (0.33360,5.87830) (0.33880,5.70890) (0.34330,4.96760)};
\addplot[color=olive!80!black, mark=x, mark size=1.45pt, line width=0.65pt] coordinates {(0.32430,5.94340) (0.32580,5.88780) (0.32990,5.67960) (0.33210,5.27240)};
\addplot[color=teal!80!black, mark=square*, mark size=1.45pt, line width=0.65pt] coordinates {(0.33290,5.91880) (0.34330,5.81800) (0.35950,5.57720) (0.36930,5.41130)};
\addplot[color=orange!85!black, mark=triangle, mark size=1.45pt, line width=0.65pt] coordinates {(0.32390,5.94460) (0.32820,5.86200) (0.32900,5.78640) (0.33430,5.57420)};
\addplot[color=violet!85!black, mark=pentagon*, mark size=1.45pt, line width=0.65pt] coordinates {(0.32400,5.88120) (0.33380,5.83190) (0.35420,5.59200) (0.37500,5.27730)};
\addplot[color=blue!70!black, mark=+, mark size=1.45pt, line width=0.65pt] coordinates {(0.35060,5.95820) (0.37790,5.70410) (0.39690,5.51910) (0.40880,5.46800)};
\addplot[only marks, draw=black, fill=yellow, mark=filled pentastar, mark size=3pt] coordinates {(0.32120,5.92340)};
\end{axis}
\end{tikzpicture}
\end{subfigure}
\vspace{1mm}
\begin{tikzpicture}[font=\fontfamily{phv}\selectfont]
\draw[draw=black, fill=yellow] plot[mark=filled pentastar, mark size=3pt] coordinates {(0.185,0.00)};
\node[anchor=west, font=\tiny] at (0.46,0.00) {Baseline};
\draw[color=red, line width=0.65pt] (2.30,0.00) -- (2.67,0.00);
\draw[color=red, line width=0.65pt] plot[mark=*, mark size=1.45pt] coordinates {(2.485,0.00)};
\node[anchor=west, font=\tiny] at (2.76,0.00) {Ours};
\draw[color=magenta!75!black, line width=0.65pt] (4.60,0.00) -- (4.97,0.00);
\draw[color=magenta!75!black, line width=0.65pt] plot[mark=diamond*, mark size=1.45pt] coordinates {(4.785,0.00)};
\node[anchor=west, font=\tiny] at (5.06,0.00) {NegToMe};
\draw[color=gray!85!black, line width=0.65pt] (6.90,0.00) -- (7.27,0.00);
\draw[color=gray!85!black, line width=0.65pt] plot[mark=star, mark size=1.45pt] coordinates {(7.085,0.00)};
\node[anchor=west, font=\tiny] at (7.36,0.00) {SPARK};
\draw[color=olive!80!black, line width=0.65pt] (9.20,0.00) -- (9.57,0.00);
\draw[color=olive!80!black, line width=0.65pt] plot[mark=x, mark size=1.45pt] coordinates {(9.385,0.00)};
\node[anchor=west, font=\tiny] at (9.66,0.00) {PG};
\draw[color=teal!80!black, line width=0.65pt] (0.00,-0.38) -- (0.37,-0.38);
\draw[color=teal!80!black, line width=0.65pt] plot[mark=square*, mark size=1.45pt] coordinates {(0.185,-0.38)};
\node[anchor=west, font=\tiny] at (0.46,-0.38) {CADS};
\draw[color=cyan!65!black, line width=0.65pt] (2.30,-0.38) -- (2.67,-0.38);
\draw[color=cyan!65!black, line width=0.65pt] plot[mark=triangle*, mark size=1.45pt] coordinates {(2.485,-0.38)};
\node[anchor=west, font=\tiny] at (2.76,-0.38) {Contextual};
\draw[color=orange!85!black, line width=0.65pt] (4.60,-0.38) -- (4.97,-0.38);
\draw[color=orange!85!black, line width=0.65pt] plot[mark=triangle, mark size=1.45pt] coordinates {(4.785,-0.38)};
\node[anchor=west, font=\tiny] at (5.06,-0.38) {Shield};
\draw[color=violet!85!black, line width=0.65pt] (6.90,-0.38) -- (7.27,-0.38);
\draw[color=violet!85!black, line width=0.65pt] plot[mark=pentagon*, mark size=1.45pt] coordinates {(7.085,-0.38)};
\node[anchor=west, font=\tiny] at (7.36,-0.38) {CNO};
\draw[color=blue!70!black, line width=0.65pt] (9.20,-0.38) -- (9.57,-0.38);
\draw[color=blue!70!black, line width=0.65pt] plot[mark=+, mark size=1.45pt] coordinates {(9.385,-0.38)};
\node[anchor=west, font=\tiny] at (9.66,-0.38) {Group};
\end{tikzpicture}
\caption{Diversity--quality trade-offs on GenEval across three models. 
Each curve varies one method-specific hyperparameter; the yellow star marks the model baseline.}
\label{fig:geneval_swap_comparison}
\end{figure}

\begin{table}[t]
\centering
\caption{Single-setting comparison on GenEval at HPSv3 thresholds of 7.4, 8.8, and 5.5 for FLUX.1-dev, FLUX.2-Klein-9B, and SDXL, respectively.}
\label{tab:comparison}
\resizebox{\textwidth}{!}{%
\begin{tabular}{l c ccc ccccc}
\toprule
\multirow{2}{*}{Method} & \multirow{2}{*}{\shortstack{Latency over\\baseline$\downarrow$}} & \multicolumn{3}{c}{Diversity} & \multicolumn{5}{c}{Image Fidelity} \\
\cmidrule(lr){3-5} \cmidrule(lr){6-10}
 & & DreamSim$\uparrow$ & Vendi$\uparrow$ & LPIPS$\uparrow$ & HPSv3$\uparrow$ & CLIP$\uparrow$ & ImR$\uparrow$ & PickS$\uparrow$ & GenEval$\uparrow$ \\
\midrule
\multicolumn{10}{l}{\textit{FLUX.1-dev:}} \\
  Baseline     & 1.00$\times$ & 0.2561 & 1.8747 & 0.5821 & 7.5266 & 0.3613 & 0.9579 & \textbf{23.3245} & 0.6653 \\
  CADS         & 1.00$\times$ & 0.2599 & 1.8694 & 0.5869 & \underline{7.5498} & 0.3608 & 0.9690 & \underline{23.3233} & 0.6551 \\
  PG           & 1.00$\times$ & 0.2593 & 1.8786 & 0.6019 & 7.4796 & 0.3606 & 0.9798 & 23.2855 & 0.6598 \\
  NegToMe      & 1.01$\times$ & 0.3151 & \underline{1.9829} & \underline{0.6572} & 7.3467 & 0.3616 & \underline{1.0055} & 23.1479 & 0.6670 \\
  CNO          & 1.13$\times$ & 0.2865 & 1.9288 & 0.6416 & 7.3931 & \textbf{0.3642} & \textbf{1.0084} & 23.3135 & \underline{0.6671} \\
  Shield       & 1.00$\times$ & 0.2813 & 1.9112 & 0.6221 & 7.4127 & \underline{0.3624} & 0.9734 & 23.2872 & 0.6633 \\
  Group        & 1.54$\times$ & 0.3045 & 1.9727 & 0.6325 & 7.4173 & 0.3620 & 0.9264 & 23.2908 & \textbf{0.6723} \\
  Contextual   & 1.28$\times$ & \underline{0.3225} & 1.9793 & \textbf{0.6994} & \textbf{7.6795} & 0.3533 & 0.9296 & 23.2449 & 0.6404 \\
\cmidrule{1-10}
  Ours & 1.15$\times$ & \textbf{0.3314} & \textbf{2.0090} & 0.6519 & 7.4321 & 0.3619 & 0.9603 & 23.2617 & 0.6619 \\
\midrule
\multicolumn{10}{l}{\textit{FLUX.2-Klein-9B:}} \\
  Baseline     & 1.00$\times$ & 0.1964 & 1.6791 & 0.5495 & \textbf{9.0595} & 0.3876 & \textbf{1.3912} & \textbf{23.6764} & \textbf{0.8813} \\
  CADS         & 1.01$\times$ & 0.2258 & 1.7567 & 0.5711 & 8.9555 & 0.3859 & 1.3712 & \underline{23.6665} & 0.8324 \\
  PG           & 1.01$\times$ & 0.2031 & 1.6799 & 0.5787 & 8.9348 & 0.3855 & 1.3889 & 23.4989 & 0.8632 \\
  NegToMe      & 1.00$\times$ & 0.2426 & 1.7955 & 0.6373 & 8.8926 & 0.3857 & 1.3454 & 23.6210 & 0.8504 \\
  CNO          & 1.97$\times$ & 0.2070 & 1.7079 & 0.6032 & 8.8135 & \textbf{0.3891} & 1.3840 & 23.5130 & 0.8672 \\
  Shield       & 1.01$\times$ & 0.2386 & 1.7401 & \textbf{0.6735} & 8.8634 & \underline{0.3887} & \underline{1.3902} & 23.3728 & \underline{0.8698} \\
  Group        & 1.99$\times$ & 0.2410 & 1.7915 & 0.6157 & 8.8540 & 0.3861 & 1.3517 & 23.6074 & 0.8540 \\
  Contextual   & 3.52$\times$ & \underline{0.2519} & \underline{1.8037} & 0.6093 & 8.9387 & 0.3859 & 1.3736 & 23.6401 & 0.8573 \\
\cmidrule{1-10}
  Ours & 1.32$\times$ & \textbf{0.2643} & \textbf{1.8280} & \underline{0.6599} & \underline{8.9825} & 0.3860 & 1.3430 & 23.6246 & 0.8525 \\
\midrule

\multicolumn{10}{l}{\textit{SDXL:}} \\
  Baseline     & 1.00$\times$ & 0.3212 & 1.9565 & 0.7243 & \textbf{5.9234} & \underline{0.3828} & \underline{0.9481} & \textbf{22.9723} & 0.5539 \\
  CADS         & 1.00$\times$ & 0.3595 & 2.0310 & 0.7264 & 5.5772 & 0.3786 & 0.8433 & 22.8572 & 0.5464 \\
  SPARKE       & 1.00$\times$ & 0.3388 & 1.9795 & \textbf{0.7713} & 5.7089 & 0.3797 & 0.9398 & 22.8326 & 0.5477 \\
  PG           & 1.00$\times$ & 0.3299 & 1.9598 & 0.7531 & 5.6796 & 0.3811 & \textbf{0.9589} & 22.7121 & 0.5605 \\
  NegToMe      & 1.00$\times$ & 0.3393 & 2.0042 & 0.7459 & \underline{5.7745} & 0.3822 & 0.9311 & \underline{22.8969} & 0.5638 \\
  CNO          & 1.09$\times$ & 0.3542 & 2.0364 & 0.7587 & 5.5920 & 0.3820 & 0.9154 & 22.8427 & 0.5626 \\
  Shield       & 1.00$\times$ & 0.3343 & 1.9852 & \underline{0.7604} & 5.5742 & 0.3819 & 0.9180 & 22.7516 & 0.5587 \\
  Group        & 10.38$\times$ & \textbf{0.3969} & \textbf{2.1468} & 0.7583 & 5.5191 & \textbf{0.3838} & 0.9303 & 22.8413 & \textbf{0.5784} \\
\cmidrule{1-10}
  Ours & 1.05$\times$ & \underline{0.3700} & \underline{2.0863} & 0.7480 & 5.6233 & 0.3804 & 0.8938 & 22.7996 & \underline{0.5745} \\
\bottomrule
\end{tabular}%
}
\end{table}

\paragraph{Discussion.}
Our trade-off parameter $\delta$ directly specifies the token rotation angle, making it easier to choose consistently across models,
while other methods' trade-off parameters are less interpretable, highly sensitive, and harder to select.
The settings used to plot the trade-off curves for each method and model are listed in \cref{tab:geneval_swap_settings}.
Contextual Repulsion and Group Inference are two strong alternatives but have their own limitations.
Contextual Repulsion~\cite{contextual_repulsion} repels text-branch representations in multimodal attention and achieves high HPSv3 scores,
but shows a larger drop in text-to-image alignment, as reflected by substantial declines in CLIP and GenEval scores.
Group Inference~\cite{group_inference} is effective on SDXL, whose baseline diversity is relatively high, 
but yields much smaller diversity gains on models with concentrated distributions, such as FLUX.2-Klein-9B.
It also requires substantially more computation.
Several other methods offer weaker performance but add negligible computational overhead, 
making them attractive when efficiency is important.

\input{figures_arxiv/qualitative_geneval/qualitative_geneval_flux}
\input{figures_arxiv/qualitative_geneval/qualitative_geneval_flux2k}
\input{figures_arxiv/qualitative_geneval/qualitative_geneval_sdxl}

\subsection{Ablations and Analysis}
\label{exp:analysis}
We conduct ablation studies and analyze the design choices of TokenRepel,
including the effects of the repulsion hyperparameters, the fidelity-preserving procedure,
batch size, and the number of inference steps over which TokenRepel is applied.

\begin{figure}[t]
    \centering
    \begin{minipage}[t]{0.34\textwidth}
        \centering
        \begin{tikzpicture}[font=\fontfamily{phv}\selectfont]
\begin{axis}[
    width=5.20cm, height=4.0cm,
    label style={font=\scriptsize}, tick label style={font=\tiny},
    axis lines=left, scaled ticks=false, tick style={black!60},
    xmin=0.19, xmax=0.31, xtick={0.20,0.25,0.30},
    ymin=8.30, ymax=9.15, ytick={8.4,8.6,8.8,9.0},
    xticklabel style={/pgf/number format/fixed, /pgf/number format/precision=2},
    yticklabel style={/pgf/number format/fixed, /pgf/number format/precision=1, xshift=2pt},
    extra description/.code={%
        \node[font=\scriptsize, rotate=90, anchor=center] at (axis description cs:-0.17,0.5) {HPSv3 $\uparrow$};%
        \node[font=\scriptsize, anchor=center] at (axis description cs:0.5,-0.21) {DreamSim $\uparrow$};%
    },
    legend style={font=\tiny, at={(0.02,0.02)}, anchor=south west, draw=none, fill=none},
]
\addplot[color=blue!70, mark=*, mark size=1.6pt, line width=0.8pt] coordinates {
    (0.1964,9.0595)
    (0.2315,9.0477)
    (0.2632,9.0101)
    (0.2803,8.8281)
    (0.2977,8.3823)
};
\addlegendentry{$\delta \uparrow$}
\end{axis}
\end{tikzpicture}
    \end{minipage}
    \hspace{-1em}
    \begin{minipage}[t]{0.34\textwidth}
        \centering
        \begin{tikzpicture}[font=\fontfamily{phv}\selectfont]
\begin{axis}[
    width=5.20cm, height=4.0cm,
    label style={font=\scriptsize}, tick label style={font=\tiny},
    axis lines=left, scaled ticks=false, tick style={black!60},
    xmin=0.19, xmax=0.31, xtick={0.20,0.25,0.30},
    ymin=8.60, ymax=9.15, ytick={8.6,8.8,9.0},
    xticklabel style={/pgf/number format/fixed, /pgf/number format/precision=2},
    yticklabel style={/pgf/number format/fixed, /pgf/number format/precision=1, xshift=2pt},
    extra description/.code={%
        \node[font=\scriptsize, rotate=90, anchor=center] at (axis description cs:-0.17,0.5) {HPSv3 $\uparrow$};%
        \node[font=\scriptsize, anchor=center] at (axis description cs:0.5,-0.21) {DreamSim $\uparrow$};%
    },
    legend style={font=\tiny, at={(0.02,0.02)}, anchor=south west, draw=none, fill=none},
]
\addplot[color=orange!90!black, mark=*, mark size=1.6pt, line width=0.8pt] coordinates {
    (0.1964,9.0595)
    (0.2151,9.0719)
    (0.2546,8.8938)
    (0.2803,8.8281)
    (0.3019,8.6604)
};
\addlegendentry{$\bar{\theta} \uparrow$}
\end{axis}
\end{tikzpicture}
    \end{minipage}
    \hspace{-1em}
    \begin{minipage}[t]{0.34\textwidth}
        \centering
        \begin{tikzpicture}[font=\fontfamily{phv}\selectfont]
\begin{axis}[
    width=5.20cm, height=4.0cm,
    label style={font=\scriptsize}, tick label style={font=\tiny},
    axis lines=left, scaled ticks=false, tick style={black!60},
    xmin=0.225, xmax=0.315, xtick={0.23,0.26,0.29},
    ymin=8.58, ymax=9.03, ytick={8.6,8.8,9.0},
    xticklabel style={/pgf/number format/fixed, /pgf/number format/precision=2},
    yticklabel style={/pgf/number format/fixed, /pgf/number format/precision=1, xshift=2pt},
    extra description/.code={%
        \node[font=\scriptsize, rotate=90, anchor=center] at (axis description cs:-0.17,0.5) {HPSv3 $\uparrow$};%
        \node[font=\scriptsize, anchor=center] at (axis description cs:0.5,-0.21) {DreamSim $\uparrow$};%
    },
]
\addplot[only marks, mark=*, mark size=2.5pt, color=black!15] coordinates {(0.2338,8.9385)};
\addplot[only marks, mark=*, mark size=2.5pt, color=black!30] coordinates {(0.2686,8.7527)};
\addplot[only marks, mark=*, mark size=2.5pt, color=black!45] coordinates {(0.2979,8.6657)};
\addplot[only marks, mark=*, mark size=2.5pt, color=black!65] coordinates {(0.2834,8.7927)};
\addplot[only marks, mark=*, mark size=2.5pt, color=black!85] coordinates {(0.2803,8.8281)};
\node[font=\tiny, anchor=south west] at (axis cs:0.2338,8.9385) {$\kappa{=}1$};
\node[font=\tiny, anchor=north east] at (axis cs:0.2686,8.7527) {$\kappa{=}10$};
\node[font=\tiny, anchor=north east] at (axis cs:0.2979,8.6657) {$\kappa{=}100$};
\node[font=\tiny, anchor=south west] at (axis cs:0.2834,8.7927) {$\kappa{=}10^3$};
\node[font=\tiny, anchor=north east] at (axis cs:0.2803,8.8281) {$\kappa{=}\infty$};
\end{axis}
\end{tikzpicture}
    \end{minipage}
    \caption{Hyperparameter sweeps for TokenRepel without token density preservation on GenEval using FLUX.2-Klein-9B. 
    Left: $\delta\in\{0^\circ,10^\circ,20^\circ,30^\circ,40^\circ\}$ with $\bar{\theta}=60^\circ$. 
    Middle: $\bar{\theta}\in\{0^\circ,20^\circ,40^\circ,60^\circ,80^\circ\}$ with $\delta=30^\circ$. 
    Right: concentration $\kappa\in\{1,10,100,10^3,\infty\}$, where $\kappa=\infty$ corresponds to argmax, with $\delta=30^\circ$ and $\bar{\theta}=60^\circ$.}
    \label{fig:analysis}
\end{figure}
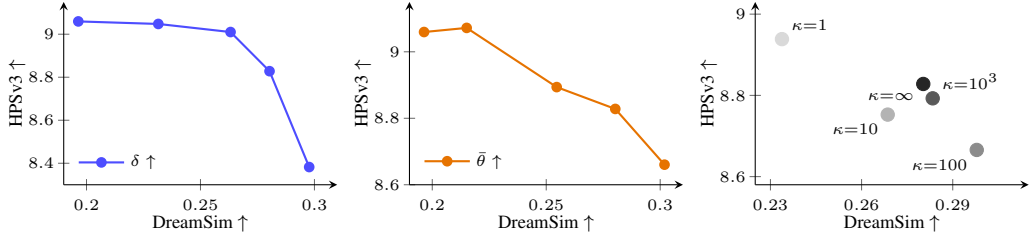

\paragraph{Effect of hyperparameters.}
We analyze the impact of the hyperparameters $\delta$ and $\bar{\theta}$.
$\delta$ controls the repulsion strength: larger $\delta$ increases the angular separation between similar features,
while $\delta=0$ reduces to the baseline.
$\bar{\theta}$ sets the angular threshold, determining which features are repelled; only pairs within this angle are affected,
while $\bar{\theta}=0$ reduces to the baseline.
We report sweeps over both parameters for our method without token density preservation in \cref{fig:analysis}.
Increasing $\delta$ and $\bar{\theta}$ both increase diversity at the cost of quality.
In addition, a higher concentration $\kappa$ in \cref{eq:vmf_softmax_mean} increases diversity with a slight drop in quality.
In this paper, we use the zero-temperature setting, which converts softmax to argmax to select the most similar token,
yielding a simpler design.

\paragraph{Ablation of token density preservation.}
We compare TokenRepel without token density preservation (w/o P), with blockwise preservation applied only to the final block output (B),
and with layerwise preservation (L) across all three models on GenEval in \cref{fig:geneval_density_constraint_comparison}.
For the FLUX models, at comparable diversity levels,
layerwise preservation generally achieves higher HPSv3 than the other variants.
For SDXL, this effect is less pronounced.
In the appendix, \cref{fig:first_n_blocks_correction} shows that correcting more layers improves fidelity but reduces diversity.
Correcting only the early layers can better balance fidelity and diversity while saving computation,
as the two auxiliary branches can stop early.

\pgfdeclareplotmark{geneval density pentastar}{%
  \pgfmathsetlengthmacro{\genevalstarinner}{0.382*\pgfplotmarksize}%
  \pgfpathmoveto{\pgfqpointpolar{90}{\pgfplotmarksize}}%
  \pgfpathlineto{\pgfqpointpolar{54}{\genevalstarinner}}%
  \pgfpathlineto{\pgfqpointpolar{18}{\pgfplotmarksize}}%
  \pgfpathlineto{\pgfqpointpolar{-18}{\genevalstarinner}}%
  \pgfpathlineto{\pgfqpointpolar{-54}{\pgfplotmarksize}}%
  \pgfpathlineto{\pgfqpointpolar{-90}{\genevalstarinner}}%
  \pgfpathlineto{\pgfqpointpolar{-126}{\pgfplotmarksize}}%
  \pgfpathlineto{\pgfqpointpolar{-162}{\genevalstarinner}}%
  \pgfpathlineto{\pgfqpointpolar{162}{\pgfplotmarksize}}%
  \pgfpathlineto{\pgfqpointpolar{126}{\genevalstarinner}}%
  \pgfpathclose\pgfusepathqfillstroke%
}
\begin{figure}[t]
\centering
\begin{subfigure}[t]{0.34\textwidth}
\centering
\begin{tikzpicture}[font=\fontfamily{phv}\selectfont]
\begin{axis}[
  width=5.20cm, height=4.0cm,
  title={FLUX.1-dev}, title style={font=\small},
  label style={font=\scriptsize}, tick label style={font=\tiny},
  axis lines=left, scaled ticks=false, tick style={black!60},
  xmin=0.25, xmax=0.365, xtick={0.26,0.30,0.34},
  ymin=7.0, ymax=7.7, ytick={7.2,7.4,7.6},
  xticklabel style={/pgf/number format/fixed, /pgf/number format/precision=2},
  yticklabel style={/pgf/number format/fixed, /pgf/number format/precision=1, xshift=2pt},
  extra description/.code={%
    \node[font=\scriptsize, rotate=90, anchor=center] at (axis description cs:-0.17,0.5) {HPSv3 $\uparrow$};%
    \node[font=\scriptsize, anchor=center] at (axis description cs:0.5,-0.21) {DreamSim $\uparrow$};%
  },
]
\addplot[color=gray!80!black, mark=square*, mark size=1.45pt, line width=0.65pt] coordinates {(0.2584,7.5656) (0.2793,7.4923) (0.3010,7.3947) (0.3176,7.3402) (0.3399,7.1667)};
\addplot[color=blue!70!black, mark=triangle*, mark size=1.45pt, line width=0.65pt] coordinates {(0.2584,7.5656) (0.2897,7.4817) (0.3156,7.3785) (0.3173,7.2404) (0.3480,7.0928)};
\addplot[color=red, mark=*, mark size=1.45pt, line width=0.65pt] coordinates {(0.2584,7.5656) (0.2782,7.5082) (0.3002,7.4589) (0.3314,7.4321) (0.3550,7.2391)};
\addplot[only marks, draw=black, fill=yellow, mark=geneval density pentastar, mark size=3pt] coordinates {(0.2584,7.5656)};
\end{axis}
\end{tikzpicture}
\end{subfigure}
\hspace{-1em}
\begin{subfigure}[t]{0.34\textwidth}
\centering
\begin{tikzpicture}[font=\fontfamily{phv}\selectfont]
\begin{axis}[
  width=5.20cm, height=4.0cm,
  title={FLUX.2-Klein-9B}, title style={font=\small},
  label style={font=\scriptsize}, tick label style={font=\tiny},
  axis lines=left, scaled ticks=false, tick style={black!60},
  xmin=0.19, xmax=0.305, xtick={0.20,0.24,0.28},
  ymin=8.6, ymax=9.2, ytick={8.8,9.0,9.2},
  xticklabel style={/pgf/number format/fixed, /pgf/number format/precision=2},
  yticklabel style={/pgf/number format/fixed, /pgf/number format/precision=1, xshift=2pt},
  extra description/.code={%
    \node[font=\scriptsize, rotate=90, anchor=center] at (axis description cs:-0.17,0.5) {HPSv3 $\uparrow$};%
    \node[font=\scriptsize, anchor=center] at (axis description cs:0.5,-0.21) {DreamSim $\uparrow$};%
  },
]
\addplot[color=gray!80!black, mark=square*, mark size=1.45pt, line width=0.65pt] coordinates {(0.1964,9.0592) (0.2315,9.0477) (0.2632,8.9201) (0.2803,8.7281) (0.2977,8.3823)};
\addplot[color=blue!70!black, mark=triangle*, mark size=1.45pt, line width=0.65pt] coordinates {(0.1964,9.0592) (0.2486,9.0947) (0.2536,8.8739) (0.2622,8.8099) (0.2616,8.8584)};
\addplot[color=red, mark=*, mark size=1.45pt, line width=0.65pt] coordinates {(0.1964,9.0592) (0.2534,9.0646) (0.2643,8.9825) (0.2799,8.7870) (0.2806,8.7634)};
\addplot[only marks, draw=black, fill=yellow, mark=geneval density pentastar, mark size=3pt] coordinates {(0.1964,9.0592)};
\end{axis}
\end{tikzpicture}
\end{subfigure}
\hspace{-1em}
\begin{subfigure}[t]{0.34\textwidth}
\centering
\begin{tikzpicture}[font=\fontfamily{phv}\selectfont]
\begin{axis}[
  width=5.20cm, height=4.0cm,
  title={SDXL}, title style={font=\small},
  label style={font=\scriptsize}, tick label style={font=\tiny},
  axis lines=left, scaled ticks=false, tick style={black!60},
  xmin=0.315, xmax=0.395, xtick={0.32,0.35,0.38},
  ymin=5.35, ymax=6.0, ytick={5.4,5.6,5.8},
  xticklabel style={/pgf/number format/fixed, /pgf/number format/precision=2},
  yticklabel style={/pgf/number format/fixed, /pgf/number format/precision=1, xshift=2pt},
  extra description/.code={%
    \node[font=\scriptsize, rotate=90, anchor=center] at (axis description cs:-0.17,0.5) {HPSv3 $\uparrow$};%
    \node[font=\scriptsize, anchor=center] at (axis description cs:0.5,-0.21) {DreamSim $\uparrow$};%
  },
]
\addplot[color=gray!80!black, mark=square*, mark size=1.45pt, line width=0.65pt] coordinates {(0.3212,5.9234) (0.3344,5.8429) (0.3544,5.7184) (0.3731,5.5896) (0.3836,5.4208)};
\addplot[color=blue!70!black, mark=triangle*, mark size=1.45pt, line width=0.65pt] coordinates {(0.3212,5.9234) (0.3409,5.8637) (0.3560,5.7321) (0.3653,5.5747) (0.3699,5.5249)};
\addplot[color=red, mark=*, mark size=1.45pt, line width=0.65pt] coordinates {(0.3212,5.9234) (0.3459,5.8322) (0.3700,5.6233) (0.3851,5.4437) (0.3892,5.4396)};
\addplot[only marks, draw=black, fill=yellow, mark=geneval density pentastar, mark size=3pt] coordinates {(0.3212,5.9234)};
\end{axis}
\end{tikzpicture}
\end{subfigure}
\vspace{1mm}
\begin{tikzpicture}[font=\fontfamily{phv}\selectfont]
\draw[draw=black, fill=yellow] plot[mark=geneval density pentastar, mark size=3pt] coordinates {(0.185,0)};
\node[anchor=west, font=\tiny] at (0.46,0) {Baseline};
\draw[color=gray!80!black, line width=0.65pt] (2.70,0) -- (3.07,0);
\draw[color=gray!80!black, line width=0.65pt] plot[mark=square*, mark size=1.45pt] coordinates {(2.885,0)};
\node[anchor=west, font=\tiny] at (3.16,0) {w/o P};
\draw[color=blue!70!black, line width=0.65pt] (5.40,0) -- (5.77,0);
\draw[color=blue!70!black, line width=0.65pt] plot[mark=triangle*, mark size=1.45pt] coordinates {(5.585,0)};
\node[anchor=west, font=\tiny] at (5.86,0) {B};
\draw[color=red, line width=0.65pt] (8.10,0) -- (8.47,0);
\draw[color=red, line width=0.65pt] plot[mark=*, mark size=1.45pt] coordinates {(8.285,0)};
\node[anchor=west, font=\tiny] at (8.56,0) {L};
\end{tikzpicture}
\caption{GenEval diversity--quality trade-offs for FLUX.1-dev, FLUX.2-Klein-9B, and SDXL. Each curve sweeps the repulsion angle $\delta$;
w/o P uses no density preservation, B applies it to the final block output,
and L applies it layerwise within the denoising network. The yellow star marks the model baseline.}
\label{fig:geneval_density_constraint_comparison}
\end{figure}
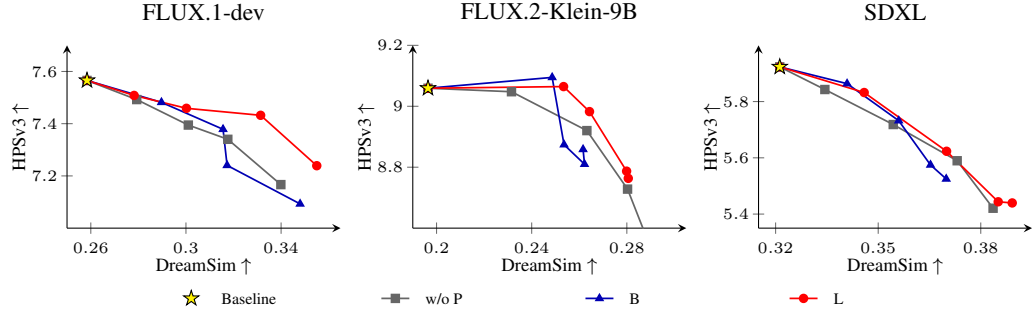

\paragraph{Batch size analysis.}
We evaluate TokenRepel with batch sizes from 2 to 12 using FLUX.2-Klein-9B and matched seeds.
As shown in \cref{fig:batch_size_analysis}, the HPSv3 gap to the baseline remains relatively stable as batch size increases,
DreamSim decreases slightly, and Vendi Score rises with the size of the generated set.
Because DreamSim is a pairwise metric, larger batches include more image pairs,
increasing the likelihood that some pairs converge to similar modes after a one-step repulsion.
In contrast, the Vendi Score increases with batch size because it estimates the effective number of distinct samples in the evaluated set.
This difference is therefore statistical and does not indicate that either larger or smaller batches produce a more diverse distribution.

\paragraph{Number of repulsion steps.}
We examine the effectiveness of TokenRepel across different repulsion windows,
varying the window from the first 3 steps to the first 24 of the 25 inference steps of FLUX.1-dev.
As shown in the left and middle panels of \cref{fig:repulsion_steps_analysis},
extending this window consistently improves diversity, but the gains quickly saturate while HPSv3 steadily decreases.
To explain this saturation, we measure the fraction of image tokens whose cross-image similarity exceeds the repulsion threshold at each timestep,
averaged across layers and prompts.
As shown in the right panel of \cref{fig:repulsion_steps_analysis},
this fraction falls from around 86\% at the first step and remains around 10\% near the end of sampling.
Early steps shape coarse global structure and contain more similar cross-image patches,
so repulsion affects more tokens.
Later steps refine details, leaving fewer similar tokens to repel.
We therefore apply repulsion only during the first few steps,
capturing most of the diversity gain while reducing the additional computation.

\subsection{Further Discussion}

\paragraph{Generalization to image editing.}
We evaluate TokenRepel on KontextBench, which contains 1,026 image--instruction pairs from 108 source images.
For evaluation, an LLM generates the target prompt from the source image and editing instruction.
We generate four images for each image--instruction pair using FLUX Kontext.
As shown in \cref{tab:kontextbench}, TokenRepel substantially improves output diversity,
with only a small drop in editing quality.
Its effectiveness depends on the instruction.
It offers limited benefit for deterministic edits, such as object removal,
but works well for open-ended edits, such as ``add a cowboy hat'' or generating content in the reference style.
\begin{table}[t]
\centering
\caption{Image-editing results on KontextBench using FLUX Kontext.}
\label{tab:kontextbench}
\small
\begin{tabular}{l ccc cccc}
\toprule
Method & DreamSim$\uparrow$ & Vendi$\uparrow$ & LPIPS$\uparrow$ & HPSv3$\uparrow$ & CLIP$\uparrow$ & ImageReward$\uparrow$ & PickScore$\uparrow$ \\
\midrule
Baseline & 0.098 & 1.425 & 0.320 & 9.384 & 0.364 & 0.582 & 21.777 \\
Ours     & 0.209 & 1.699 & 0.502 & 9.301 & 0.357 & 0.509 & 21.591 \\
\bottomrule
\end{tabular}
\end{table}

\clearpage
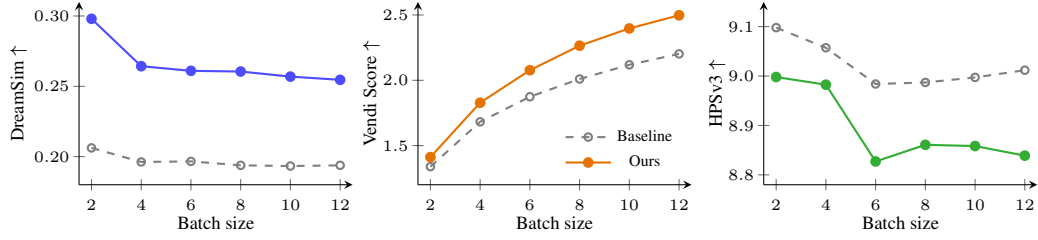
\begin{figure}[t]
    \centering
    \begin{minipage}[t]{0.34\textwidth}
        \centering
        \begin{tikzpicture}[font=\fontfamily{phv}\selectfont]
        \begin{axis}[
            width=5.20cm, height=4.0cm,
            label style={font=\scriptsize}, tick label style={font=\tiny},
            axis lines=left, scaled ticks=false,
            tick style={black!60},
            xmin=1.5, xmax=12.5, ymin=0.18, ymax=0.31,
            xtick={2,4,6,8,10,12},
            ytick={0.20,0.25,0.30},
            yticklabel style={/pgf/number format/fixed, /pgf/number format/fixed zerofill, /pgf/number format/precision=2, xshift=2pt},
            extra description/.code={%
                \node[font=\scriptsize, rotate=90, anchor=center] at (axis description cs:-0.22,0.5) {DreamSim $\uparrow$};%
                \node[font=\scriptsize, anchor=center] at (axis description cs:0.5,-0.21) {Batch size};%
            },
        ]
        \addplot[black!50, dashed, mark=o, mark options={solid}, mark size=1.4pt, line width=0.8pt] coordinates {
            (2,0.2062) (4,0.1962) (6,0.1966) (8,0.1938) (10,0.1933) (12,0.1938)
        };
        \addplot[myblue, mark=*, mark size=1.6pt, line width=0.8pt] coordinates {
            (2,0.2980) (4,0.2643) (6,0.2610) (8,0.2605) (10,0.2569) (12,0.2546)
        };
        \end{axis}
        \end{tikzpicture}
    \end{minipage}
    \hspace{-1em}
    \begin{minipage}[t]{0.34\textwidth}
        \centering
        \begin{tikzpicture}[font=\fontfamily{phv}\selectfont]
        \begin{axis}[
            width=5.20cm, height=4.0cm,
            label style={font=\scriptsize}, tick label style={font=\tiny},
            axis lines=left, scaled ticks=false,
            tick style={black!60},
            xmin=1.5, xmax=12.5, ymin=1.2, ymax=2.6,
            xtick={2,4,6,8,10,12},
            ytick={1.5,2.0,2.5},
            yticklabel style={/pgf/number format/fixed, /pgf/number format/fixed zerofill, /pgf/number format/precision=1, xshift=2pt},
            extra description/.code={%
                \node[font=\scriptsize, rotate=90, anchor=center] at (axis description cs:-0.17,0.5) {Vendi Score $\uparrow$};%
                \node[font=\scriptsize, anchor=center] at (axis description cs:0.5,-0.21) {Batch size};%
            },
            legend style={font=\tiny, at={(0.98,0.02)}, anchor=south east, draw=none, fill=none},
        ]
        \addplot[black!50, dashed, mark=o, mark options={solid}, mark size=1.4pt, line width=0.8pt] coordinates {
            (2,1.3382) (4,1.6819) (6,1.8739) (8,2.0097) (10,2.1186) (12,2.2022)
        };
        \addlegendentry{Baseline}
        \addplot[myorange, mark=*, mark size=1.6pt, line width=0.8pt] coordinates {
            (2,1.4114) (4,1.8280) (6,2.0771) (8,2.2649) (10,2.3969) (12,2.4977)
        };
        \addlegendentry{Ours}
        \end{axis}
        \end{tikzpicture}
    \end{minipage}
    \hspace{-1em}
    \begin{minipage}[t]{0.34\textwidth}
        \centering
        \begin{tikzpicture}[font=\fontfamily{phv}\selectfont]
        \begin{axis}[
            width=5.20cm, height=4.0cm,
            label style={font=\scriptsize}, tick label style={font=\tiny},
            axis lines=left, scaled ticks=false,
            tick style={black!60},
            xmin=1.5, xmax=12.5, ymin=8.78, ymax=9.15,
            xtick={2,4,6,8,10,12},
            ytick={8.8,8.9,9.0,9.1},
            yticklabel style={/pgf/number format/fixed, /pgf/number format/fixed zerofill, /pgf/number format/precision=1, xshift=2pt},
            extra description/.code={%
                \node[font=\scriptsize, rotate=90, anchor=center] at (axis description cs:-0.17,0.5) {HPSv3 $\uparrow$};%
                \node[font=\scriptsize, anchor=center] at (axis description cs:0.5,-0.21) {Batch size};%
            },
        ]
        \addplot[black!50, dashed, mark=o, mark options={solid}, mark size=1.4pt, line width=0.8pt] coordinates {
            (2,9.0980) (4,9.0572) (6,8.9839) (8,8.9870) (10,8.9972) (12,9.0118)
        };
        \addplot[mygreen, mark=*, mark size=1.6pt, line width=0.8pt] coordinates {
            (2,8.9980) (4,8.9825) (6,8.8271) (8,8.8608) (10,8.8583) (12,8.8389)
        };
        \end{axis}
        \end{tikzpicture}
    \end{minipage}
    \caption{Effect of batch size on DreamSim (left), Vendi Score (middle), and HPSv3 (right) for the baseline and TokenRepel with FLUX.2-Klein-9B on GenEval.}
    \label{fig:batch_size_analysis}
\end{figure}
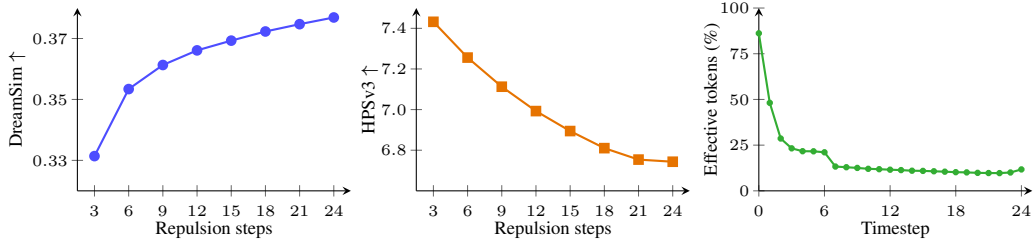
\begin{figure}[t]
    \centering
    \begin{minipage}[t]{0.34\textwidth}
        \centering
        \begin{tikzpicture}[font=\fontfamily{phv}\selectfont]
        \begin{axis}[
            width=5.20cm, height=4.0cm,
            label style={font=\scriptsize}, tick label style={font=\tiny},
            axis lines=left, scaled ticks=false,
            tick style={black!60},
            xmin=1.5, xmax=25.5, ymin=0.32, ymax=0.38,
            xtick={3,6,9,12,15,18,21,24},
            ytick={0.33,0.35,0.37},
            yticklabel style={/pgf/number format/fixed, /pgf/number format/fixed zerofill, /pgf/number format/precision=2, xshift=2pt},
            extra description/.code={%
                \node[font=\scriptsize, rotate=90, anchor=center] at (axis description cs:-0.22,0.5) {DreamSim $\uparrow$};%
                \node[font=\scriptsize, anchor=center] at (axis description cs:0.5,-0.21) {Repulsion steps};%
            },
        ]
        \addplot[myblue, mark=*, mark size=1.6pt, line width=0.8pt] coordinates {
            (3,0.3314) (6,0.3534) (9,0.3613) (12,0.3661) (15,0.3693) (18,0.3723) (21,0.3747) (24,0.3769)
        };
        \end{axis}
        \end{tikzpicture}
    \end{minipage}
    \hspace{-1em}
    \begin{minipage}[t]{0.34\textwidth}
        \centering
        \begin{tikzpicture}[font=\fontfamily{phv}\selectfont]
        \begin{axis}[
            width=5.20cm, height=4.0cm,
            label style={font=\scriptsize}, tick label style={font=\tiny},
            axis lines=left, scaled ticks=false,
            tick style={black!60},
            xmin=1.5, xmax=25.5, ymin=6.6, ymax=7.5,
            xtick={3,6,9,12,15,18,21,24},
            ytick={6.8,7.0,7.2,7.4},
            yticklabel style={/pgf/number format/fixed, /pgf/number format/fixed zerofill, /pgf/number format/precision=1, xshift=2pt},
            extra description/.code={%
                \node[font=\scriptsize, rotate=90, anchor=center] at (axis description cs:-0.17,0.5) {HPSv3 $\uparrow$};%
                \node[font=\scriptsize, anchor=center] at (axis description cs:0.5,-0.21) {Repulsion steps};%
            },
        ]
        \addplot[myorange, mark=square*, mark size=1.6pt, line width=0.8pt] coordinates {
            (3,7.4321) (6,7.2560) (9,7.1126) (12,6.9926) (15,6.8943) (18,6.8102) (21,6.7538) (24,6.7433)
        };
        \end{axis}
        \end{tikzpicture}
    \end{minipage}
    \hspace{-1em}
    \begin{minipage}[t]{0.34\textwidth}
        \centering
        \begin{tikzpicture}[font=\fontfamily{phv}\selectfont]
        \begin{axis}[
            width=5.20cm, height=4.0cm,
            label style={font=\scriptsize}, tick label style={font=\tiny},
            axis lines=left, scaled ticks=false,
            tick style={black!60},
            xmin=0, xmax=25, ymin=0, ymax=100,
            xtick={0,6,12,18,24},
            ytick={0,25,50,75,100},
            yticklabel style={xshift=2pt},
            extra description/.code={%
                \node[font=\scriptsize, rotate=90, anchor=center] at (axis description cs:-0.17,0.5) {Effective tokens (\%)};%
                \node[font=\scriptsize, anchor=center] at (axis description cs:0.5,-0.21) {Timestep};%
            },
        ]
        \addplot[mygreen, mark=*, mark size=0.8pt, line width=0.8pt] coordinates {
            (0,86.18) (1,48.14) (2,28.63) (3,23.25) (4,21.68) (5,21.63) (6,21.09) (7,13.29) (8,12.92) (9,12.54) (10,12.05) (11,11.84) (12,11.56) (13,11.35) (14,11.01) (15,10.91) (16,10.71) (17,10.48) (18,10.21) (19,10.09) (20,9.84) (21,9.74) (22,9.70) (23,10.04) (24,11.72)
        };
        \end{axis}
        \end{tikzpicture}
    \end{minipage}
    \caption{Effect of the number of repulsion steps (from 3 to 24) on DreamSim (left) and HPSv3 (middle),
    and the fraction of tokens eligible for repulsion at each timestep with $\bar{\theta}=45^\circ$ (right),
    measured with FLUX.1-dev on GenEval.}
    \label{fig:repulsion_steps_analysis}
\end{figure}
\paragraph{Limitations.}
Although our method enhances diversity, the upper bound of achievable diversity is still fundamentally limited by the support of the training data distribution learned by the model. For example, our method is significantly more effective on the FLUX models than on SDXL.
Quality preservation also varies across metrics (\cref{tab:comparison}):
it is more pronounced for HPSv3 but less effective for GenEval and PickScore.
In addition, the proposed quality-preserving mechanism is not directly applicable to autoregressive image generative models \cite{infinity},
since their predictions correspond to probabilities rather than gradient-like score estimates.

\section{Conclusion}
We have presented TokenRepel, a test-time approach for enhancing the visual diversity of generated image sets under identical prompts.
First, we outlined a token repulsion mechanism for encouraging patch-wise image diversity by taking a negative Riemannian gradient step on a distribution of features on the unit hypersphere.
Second, we showed how to correct these perturbations so that they remain in the high-density region learned by the generative model and 
extend this strategy within the transformer layers.
Our approach demonstrates substantial improvements in diversity of the output images across three T2I models,
while largely preserving the average sample fidelity.

\paragraph{Acknowledgements.}
We thank Xingjian Leng, Jaskirat Singh, and Zhengyang Yu for valuable discussions.
Dr Campbell is the recipient of an Australian Research Council Discovery Early Career Award (project number DE250100542) funded by the Australian Government.

\clearpage
{\small
\bibliographystyle{unsrtnat}
\bibliography{ref}

\begin{thebibliography}{48}
\providecommand{\natexlab}[1]{#1}
\providecommand{\url}[1]{\texttt{#1}}
\expandafter\ifx\csname urlstyle\endcsname\relax
  \providecommand{\doi}[1]{doi: #1}\else
  \providecommand{\doi}{doi: \begingroup \urlstyle{rm}\Url}\fi

\bibitem[{B. F. Labs}(2024)]{flux}
{B. F. Labs}.
\newblock {Flux.1 [dev]}.
\newblock \url{https://blackforestlabs.ai/}, 2024.

\bibitem[Rombach et~al.(2022)Rombach, Blattmann, Lorenz, Esser, and Ommer]{sd}
Robin Rombach, Andreas Blattmann, Dominik Lorenz, Patrick Esser, and Bj{\"o}rn Ommer.
\newblock High-resolution image synthesis with latent diffusion models.
\newblock In \emph{IEEE/CVF Conference on Computer Vision and Pattern Recognition}, 2022.

\bibitem[Esser et~al.(2024)Esser, Kulal, Blattmann, Entezari, M{\"u}ller, Saini, Levi, Lorenz, Sauer, Boesel, et~al.]{sd3}
Patrick Esser, Sumith Kulal, Andreas Blattmann, Rahim Entezari, Jonas M{\"u}ller, Harry Saini, Yam Levi, Dominik Lorenz, Axel Sauer, Frederic Boesel, et~al.
\newblock Scaling rectified flow transformers for high-resolution image synthesis.
\newblock In \emph{International Conference on Machine Learning}, 2024.

\bibitem[Podell et~al.(2024)Podell, English, Lacey, Blattmann, Dockhorn, M{\"u}ller, Penna, and Rombach]{sdxl}
Dustin Podell, Zion English, Kyle Lacey, Andreas Blattmann, Tim Dockhorn, Jonas M{\"u}ller, Joe Penna, and Robin Rombach.
\newblock Sdxl: Improving latent diffusion models for high-resolution image synthesis.
\newblock In \emph{International Conference on Learning Representations}, 2024.

\bibitem[Wu et~al.(2025)Wu, Li, Zhou, Lin, Gao, Yan, Yin, Bai, Xu, Chen, et~al.]{qwenimage}
Chenfei Wu, Jiahao Li, Jingren Zhou, Junyang Lin, Kaiyuan Gao, Kun Yan, Sheng-ming Yin, Shuai Bai, Xiao Xu, Yilei Chen, et~al.
\newblock Qwen-image technical report.
\newblock \emph{arXiv preprint arXiv:2508.02324}, 2025.

\bibitem[Han et~al.(2025)Han, Liu, Jiang, Yan, Zhang, Yuan, Peng, and Liu]{infinity}
Jian Han, Jinlai Liu, Yi~Jiang, Bin Yan, Yuqi Zhang, Zehuan Yuan, Bingyue Peng, and Xiaobing Liu.
\newblock Infinity: Scaling bitwise autoregressive modeling for high-resolution image synthesis.
\newblock In \emph{IEEE/CVF Conference on Computer Vision and Pattern Recognition}, 2025.

\bibitem[Ho and Salimans(2021)]{cfg}
Jonathan Ho and Tim Salimans.
\newblock Classifier-free diffusion guidance.
\newblock In \emph{NeurIPS Workshop on Deep Generative Models and Downstream Applications}, 2021.

\bibitem[Liu et~al.(2025)Liu, Huang, Wang, Liu, Li, and Ji]{diversegrpo}
Henglin Liu, Huijuan Huang, Jing Wang, Chang Liu, Xiu Li, and Xiangyang Ji.
\newblock Diversegrpo: Mitigating mode collapse in image generation via diversity-aware grpo.
\newblock \emph{arXiv preprint arXiv:2512.21514}, 2025.

\bibitem[Wang et~al.(2025)Wang, Liang, Liu, Liu, Liu, Zheng, Pang, Ma, Xie, Wang, et~al.]{grpo_guard}
Jing Wang, Jiajun Liang, Jie Liu, Henglin Liu, Gongye Liu, Jun Zheng, Wanyuan Pang, Ao~Ma, Zhenyu Xie, Xintao Wang, et~al.
\newblock Grpo-guard: Mitigating implicit over-optimization in flow matching via regulated clipping.
\newblock \emph{arXiv preprint arXiv:2510.22319}, 2025.

\bibitem[Zhu et~al.(2026)Zhu, Newman, Lutzeyer, Romero-Soriano, Drozdzal, and Russakovsky]{gass}
Ye~Zhu, Kaleb~S Newman, Johannes~F Lutzeyer, Adriana Romero-Soriano, Michal Drozdzal, and Olga Russakovsky.
\newblock Gass: Geometry-aware spherical sampling for disentangled diversity enhancement in text-to-image generation.
\newblock \emph{arXiv preprint arXiv:2602.17200}, 2026.

\bibitem[Sadat et~al.(2024)Sadat, Buhmann, Bradley, Hilliges, and Weber]{cads}
Seyedmorteza Sadat, Jakob Buhmann, Derek Bradley, Otmar Hilliges, and Romann Weber.
\newblock Cads: Unleashing the diversity of diffusion models through condition-annealed sampling.
\newblock In \emph{International Conference on Learning Representations}, 2024.

\bibitem[Karras et~al.(2023)Karras, Holynski, Wang, and Kemelmacher-Shlizerman]{dreampose}
Johanna Karras, Aleksander Holynski, Ting-Chun Wang, and Ira Kemelmacher-Shlizerman.
\newblock Dreampose: Fashion image-to-video synthesis via stable diffusion.
\newblock In \emph{IEEE/CVF International Conference on Computer Vision}, 2023.

\bibitem[Chen et~al.(2024)Chen, Yan, and Zhu]{gendataargu}
Yunhao Chen, Zihui Yan, and Yunjie Zhu.
\newblock A comprehensive survey for generative data augmentation.
\newblock \emph{Neurocomputing}, 2024.

\bibitem[Ma et~al.(2025{\natexlab{a}})Ma, Tong, Jia, Hu, Su, Zhang, Yang, Li, Jaakkola, Jia, and Xie]{tts_ma}
Nanye Ma, Shangyuan Tong, Haolin Jia, Hexiang Hu, Yu-Chuan Su, Mingda Zhang, Xuan Yang, Yandong Li, Tommi Jaakkola, Xuhui Jia, and Saining Xie.
\newblock Inference-time scaling for diffusion models beyond scaling denoising steps.
\newblock In \emph{IEEE/CVF Conference on Computer Vision and Pattern Recognition}, 2025{\natexlab{a}}.

\bibitem[Uehara et~al.(2025)Uehara, Zhao, Wang, Li, Regev, Levine, and Biancalani]{tts_tut}
Masatoshi Uehara, Yulai Zhao, Chenyu Wang, Xiner Li, Aviv Regev, Sergey Levine, and Tommaso Biancalani.
\newblock Inference-time alignment in diffusion models with reward-guided generation: Tutorial and review.
\newblock \emph{arXiv preprint arXiv:2501.09685}, 2025.

\bibitem[Um and Ye(2025)]{minority}
Soobin Um and Jong~Chul Ye.
\newblock Minority-focused text-to-image generation via prompt optimization.
\newblock In \emph{IEEE/CVF Conference on Computer Vision and Pattern Recognition}, 2025.

\bibitem[Shi et~al.(2025)Shi, Wu, Zhang, Zhang, Tao, and Qu]{closerlook}
Lianghe Shi, Meng Wu, Huijie Zhang, Zekai Zhang, Molei Tao, and Qing Qu.
\newblock A closer look at model collapse: From a generalization-to-memorization perspective.
\newblock In \emph{Advances in Neural Information Processing Systems}, 2025.

\bibitem[Datta et~al.(2024)Datta, Ku, Ramachandran, and Anderson]{prompt_expansion}
Siddhartha Datta, Alexander Ku, Deepak Ramachandran, and Peter Anderson.
\newblock Prompt expansion for adaptive text-to-image generation.
\newblock In \emph{Annual Meeting of the Association for Computational Linguistics}, 2024.

\bibitem[Harrington et~al.(2026)Harrington, Koepke, Karthik, Darrell, and Efros]{noise_opt_diverse}
Anne Harrington, A~Koepke, Shyamgopal Karthik, Trevor Darrell, and Alexei~A Efros.
\newblock It's never too late: Noise optimization for collapse recovery in trained diffusion models.
\newblock In \emph{IEEE/CVF Conference on Computer Vision and Pattern Recognition}, 2026.

\bibitem[Kim et~al.(2026{\natexlab{a}})Kim, Um, and Ye]{diverse_contrat_noise}
Byungjun Kim, Soobin Um, and Jong~Chul Ye.
\newblock Diverse text-to-image generation via contrastive noise optimization.
\newblock In \emph{International Conference on Learning Representations}, 2026{\natexlab{a}}.

\bibitem[Jalali et~al.(2025)Jalali, Haoyu, Gohari, and Farnia]{sparke}
Mohammad Jalali, LEI Haoyu, Amin Gohari, and Farzan Farnia.
\newblock Sparke: Scalable prompt-aware diversity and novelty guidance in diffusion models via rke score.
\newblock In \emph{Advances in Neural Information Processing Systems}, 2025.

\bibitem[Corso et~al.(2024)Corso, Xu, De~Bortoli, Barzilay, and Jaakkola]{particle_g}
Gabriele Corso, Yilun Xu, Valentin De~Bortoli, Regina Barzilay, and Tommi~S Jaakkola.
\newblock Particle guidance: non-iid diverse sampling with diffusion models.
\newblock In \emph{International Conference on Learning Representations}, 2024.

\bibitem[Bhat et~al.(2026)Bhat, Parihar, Bhat, and Babu]{dsm}
Pradhaan~S. Bhat, Rishubh Parihar, Abhijnya Bhat, and R.~Venkatesh Babu.
\newblock Don't settle at the mode! mitigating diversity collapse in pretrained flow models via feature self-guidance.
\newblock In \emph{European Conference on Computer Vision}, 2026.

\bibitem[Dahary et~al.(2026)Dahary, Koren, Garibi, and Cohen-Or]{contextual_repulsion}
Omer Dahary, Benaya Koren, Daniel Garibi, and Daniel Cohen-Or.
\newblock On-the-fly repulsion in the contextual space for rich diversity in diffusion transformers.
\newblock In \emph{ACM SIGGRAPH Conference Papers}, 2026.

\bibitem[Singh et~al.(2024)Singh, Li, Shi, Krishna, Choi, Koh, Cohen, Gould, Zheng, and Zettlemoyer]{negtome}
Jaskirat Singh, Lindsey Li, Weijia Shi, Ranjay Krishna, Yejin Choi, Pang~Wei Koh, Michael~F Cohen, Stephen Gould, Liang Zheng, and Luke Zettlemoyer.
\newblock Negative token merging: Image-based adversarial feature guidance.
\newblock \emph{arXiv preprint arXiv:2412.01339}, 2024.

\bibitem[{Black Forest Labs}(2026)]{flux2klein}
{Black Forest Labs}.
\newblock Flux.2 [klein].
\newblock \url{https://bfl.ai/models/flux-2-klein}, 2026.

\bibitem[Goodfellow et~al.(2020)Goodfellow, Pouget-Abadie, Mirza, Xu, Warde-Farley, Ozair, Courville, and Bengio]{gan}
Ian Goodfellow, Jean Pouget-Abadie, Mehdi Mirza, Bing Xu, David Warde-Farley, Sherjil Ozair, Aaron Courville, and Yoshua Bengio.
\newblock Generative adversarial networks.
\newblock \emph{Communications of the ACM}, 2020.

\bibitem[Dhariwal and Nichol(2021)]{diffusionbeat}
Prafulla Dhariwal and Alexander Nichol.
\newblock Diffusion models beat gans on image synthesis.
\newblock In \emph{Advances in Neural Information Processing Systems}, 2021.

\bibitem[Lipman et~al.(2023)Lipman, Chen, Ben-Hamu, Nickel, and Le]{flow_gen}
Yaron Lipman, Ricky~TQ Chen, Heli Ben-Hamu, Maximilian Nickel, and Matthew Le.
\newblock Flow matching for generative modeling.
\newblock In \emph{International Conference on Learning Representations}, 2023.

\bibitem[Sorokin et~al.(2025)Sorokin, Nakhodnov, Kuznetsov, and Alanov]{imagerefl}
Dmitrii Sorokin, Maksim Nakhodnov, Andrey Kuznetsov, and Aibek Alanov.
\newblock Imagerefl: Balancing quality and diversity in human-aligned diffusion models.
\newblock \emph{arXiv preprint arXiv:2505.22569}, 2025.

\bibitem[Gandikota and Bau(2026)]{distill_lose_diverse}
Rohit Gandikota and David Bau.
\newblock Distilling diversity and control in diffusion models.
\newblock In \emph{IEEE/CVF Winter Conference on Applications of Computer Vision}, 2026.

\bibitem[Parmar et~al.(2026)Parmar, Patashnik, Ostashev, Wang, Aberman, Narasimhan, and Zhu]{group_inference}
Gaurav Parmar, Or~Patashnik, Daniil Ostashev, Kuan-Chieh Wang, Kfir Aberman, Srinivasa Narasimhan, and Jun-Yan Zhu.
\newblock Scaling group inference for diverse and high-quality generation.
\newblock In \emph{International Conference on Learning Representations}, 2026.

\bibitem[Kim et~al.(2026{\natexlab{b}})Kim, Um, and Ye]{cno}
Byungjun Kim, Soobin Um, and Jong~Chul Ye.
\newblock Diverse text-to-image generation via contrastive noise optimization.
\newblock In \emph{International Conference on Learning Representations}, 2026{\natexlab{b}}.

\bibitem[Kirchhof et~al.(2025)Kirchhof, Thornton, B{\'e}thune, Ablin, Ndiaye, and Cuturi]{shielded_diffusion}
Michael Kirchhof, James Thornton, Louis B{\'e}thune, Pierre Ablin, Eugene Ndiaye, and Marco Cuturi.
\newblock Shielded diffusion: Generating novel and diverse images using sparse repellency.
\newblock In \emph{International Conference on Machine Learning}, 2025.

\bibitem[Pandey et~al.(2025)Pandey, Hold-Geoffroy, Gadelha, Mitra, Singh, and Guerrero]{motion_modes}
Karran Pandey, Yannick Hold-Geoffroy, Matheus Gadelha, Niloy~J. Mitra, Karan Singh, and Paul Guerrero.
\newblock Motion modes: What could happen next?
\newblock In \emph{IEEE/CVF Conference on Computer Vision and Pattern Recognition}, 2025.

\bibitem[Shen et~al.(2024)Shen, Liu, Sun, Ye, Li, Zhang, and Cao]{dreammover}
Liao Shen, Tianqi Liu, Huiqiang Sun, Xinyi Ye, Baopu Li, Jianming Zhang, and Zhiguo Cao.
\newblock Dreammover: Leveraging the prior of diffusion models for image interpolation with large motion.
\newblock In \emph{European Conference on Computer Vision}, 2024.

\bibitem[Tang et~al.(2023)Tang, Jia, Wang, Phoo, and Hariharan]{corres_diffusion}
Luming Tang, Menglin Jia, Qianqian Wang, Cheng~Perng Phoo, and Bharath Hariharan.
\newblock Emergent correspondence from image diffusion.
\newblock In \emph{Advances in Neural Information Processing Systems}, 2023.

\bibitem[Yu et~al.(2025)Yu, Singh, Yang, Tu, Zhang, Li, Hartley, and Campbell]{density_geodesics}
Qingtao Yu, Jaskirat Singh, Zhaoyuan Yang, Peter~Henry Tu, Jing Zhang, Hongdong Li, Richard Hartley, and Dylan Campbell.
\newblock Probability density geodesics in image diffusion latent space.
\newblock In \emph{IEEE/CVF Conference on Computer Vision and Pattern Recognition}, 2025.

\bibitem[Song et~al.(2021)Song, Sohl-Dickstein, Kingma, Kumar, Ermon, and Poole]{score_sde}
Yang Song, Jascha Sohl-Dickstein, Diederik~P. Kingma, Abhishek Kumar, Stefano Ermon, and Ben Poole.
\newblock Score-based generative modeling through stochastic differential equations.
\newblock In \emph{International Conference on Learning Representations}, 2021.

\bibitem[Song and Ermon(2019)]{ncsn}
Yang Song and Stefano Ermon.
\newblock Generative modeling by estimating gradients of the data distribution.
\newblock In \emph{Advances in Neural Information Processing Systems}, 2019.

\bibitem[Ghosh et~al.(2023)Ghosh, Hajishirzi, and Schmidt]{geneval}
Dhruba Ghosh, Hannaneh Hajishirzi, and Ludwig Schmidt.
\newblock Geneval: An object-focused framework for evaluating text-to-image alignment.
\newblock In \emph{Advances in Neural Information Processing Systems}, 2023.

\bibitem[Fu et~al.(2023)Fu, Tamir, Sundaram, Chai, Zhang, Dekel, and Isola]{dreamsim}
Stephanie Fu, Netanel~Y Tamir, Shobhita Sundaram, Lucy Chai, Richard Zhang, Tali Dekel, and Phillip Isola.
\newblock Dreamsim: Learning new dimensions of human visual similarity using synthetic data.
\newblock In \emph{Advances in Neural Information Processing Systems}, 2023.

\bibitem[Friedman and Dieng(2023)]{vendi}
Dan Friedman and Adji~Bousso Dieng.
\newblock The vendi score: A diversity evaluation metric for machine learning.
\newblock \emph{Transactions on Machine Learning Research}, 2023.

\bibitem[Zhang et~al.(2018)Zhang, Isola, Efros, Shechtman, and Wang]{lpips}
Richard Zhang, Phillip Isola, Alexei~A Efros, Eli Shechtman, and Oliver Wang.
\newblock The unreasonable effectiveness of deep features as a perceptual metric.
\newblock In \emph{IEEE/CVF Conference on Computer Vision and Pattern Recognition}, 2018.

\bibitem[Ma et~al.(2025{\natexlab{b}})Ma, Wu, Sun, and Li]{hpsv3}
Yuhang Ma, Xiaoshi Wu, Keqiang Sun, and Hongsheng Li.
\newblock Hpsv3: Towards wide-spectrum human preference score.
\newblock In \emph{IEEE/CVF International Conference on Computer Vision}, 2025{\natexlab{b}}.

\bibitem[Xu et~al.(2023)Xu, Liu, Wu, Tong, Li, Ding, Tang, and Dong]{imagereward}
Jiazheng Xu, Xiao Liu, Yuchen Wu, Yuxuan Tong, Qinkai Li, Ming Ding, Jie Tang, and Yuxiao Dong.
\newblock Imagereward: Learning and evaluating human preferences for text-to-image generation.
\newblock In \emph{Advances in Neural Information Processing Systems}, 2023.

\bibitem[Radford et~al.(2021)Radford, Kim, Hallacy, Ramesh, Goh, Agarwal, Sastry, Askell, Mishkin, Clark, et~al.]{clip}
Alec Radford, Jong~Wook Kim, Chris Hallacy, Aditya Ramesh, Gabriel Goh, Sandhini Agarwal, Girish Sastry, Amanda Askell, Pamela Mishkin, Jack Clark, et~al.
\newblock Learning transferable visual models from natural language supervision.
\newblock In \emph{International Conference on Machine Learning}, 2021.

\bibitem[Kirstain et~al.(2023)Kirstain, Polyak, Singer, Matiana, Penna, and Levy]{pickapic}
Yuval Kirstain, Adam Polyak, Uriel Singer, Shahbuland Matiana, Joe Penna, and Omer Levy.
\newblock Pick-a-pic: An open dataset of user preferences for text-to-image generation.
\newblock In \emph{Advances in Neural Information Processing Systems}, 2023.

\end{thebibliography}
}

\appendix
\newpage
\vbox{%
    \hsize\textwidth
    \linewidth\hsize
    \vskip 0.1in
    \hrule height 4pt
    \vskip 0.25in
    \vskip -\parskip
    \centering
    {\LARGE\bf Appendix: TokenRepel: Generating Diverse\\Image Sets Without Compromising Quality\par}
    \vskip 0.29in
    \vskip -\parskip
    \hrule height 1pt
    \vskip 0.09in
}
\suppressfloats[t]
\vskip 0.2in

\section{Additional Results}
\subsection{Full GenEval Sweep Results}
\label{app:geneval_sweep}
\Cref{tab:geneval_sweep_flux,tab:geneval_sweep_flux2k,tab:geneval_sweep_sdxl} report all metrics for every sweep point in \cref{fig:geneval_swap_comparison},
evaluated on GenEval.
\begin{table}[t]
\centering
\scriptsize
\setlength{\tabcolsep}{4pt}
\caption{All GenEval sweep results of \cref{fig:geneval_swap_comparison} for FLUX.1-dev. Bold settings are used in \cref{tab:comparison}.}
\label{tab:geneval_sweep_flux}
\begin{tabular}{@{}llr ccccccc@{}}
\toprule
Method & Parameter & Value & DreamSim$\uparrow$ & Vendi$\uparrow$ & LPIPS$\uparrow$ & HPSv3$\uparrow$ & CLIP$\uparrow$ & ImageReward$\uparrow$ & PickScore$\uparrow$ \\
\midrule
Baseline & -- & -- & 0.2561 & 1.8747 & 0.5821 & 7.5266 & 0.3613 & 0.9579 & 23.3245 \\
\midrule
Ours & $\delta$ & 5 & 0.2782 & 1.9204 & 0.6066 & 7.5045 & 0.3613 & 0.9562 & 23.3084 \\
 &  & 10 & 0.3002 & 1.9501 & 0.6271 & 7.4389 & 0.3607 & 0.9477 & 23.2911 \\
 &  & \textbf{20} & 0.3314 & 2.0090 & 0.6519 & 7.4321 & 0.3619 & 0.9603 & 23.2617 \\
 &  & 30 & 0.3550 & 2.0420 & 0.6635 & 7.2391 & 0.3606 & 0.9419 & 23.1613 \\
\midrule
NegToMe & $\alpha$ & \textbf{0.5} & 0.3151 & 1.9829 & 0.6572 & 7.3467 & 0.3616 & 1.0055 & 23.1479 \\
 &  & 0.75 & 0.3354 & 2.0126 & 0.6708 & 7.0417 & 0.3592 & 0.9692 & 23.0681 \\
 &  & 1 & 0.3525 & 2.0617 & 0.6809 & 6.8275 & 0.3592 & 0.9308 & 22.9745 \\
 &  & 1.45 & 0.3727 & 2.0930 & 0.6878 & 6.5173 & 0.3563 & 0.8936 & 22.7927 \\
\midrule
PG & coefficient & \textbf{0.05} & 0.2593 & 1.8786 & 0.6019 & 7.4796 & 0.3606 & 0.9798 & 23.2855 \\
 &  & 0.1 & 0.2623 & 1.8806 & 0.6349 & 7.3756 & 0.3575 & 0.9732 & 23.1865 \\
 &  & 0.14 & 0.2646 & 1.8838 & 0.6602 & 7.1406 & 0.3534 & 0.9427 & 23.0396 \\
 &  & 0.19 & 0.2656 & 1.8942 & 0.6795 & 6.6915 & 0.3503 & 0.8953 & 22.8370 \\
\midrule
CADS & noise scale & \textbf{0.0005} & 0.2599 & 1.8694 & 0.5869 & 7.5498 & 0.3608 & 0.9690 & 23.3233 \\
 &  & 0.001 & 0.2686 & 1.8878 & 0.6005 & 7.3399 & 0.3574 & 0.9013 & 23.2874 \\
 &  & 0.0013 & 0.2832 & 1.9118 & 0.6162 & 7.0386 & 0.3510 & 0.7526 & 23.1988 \\
 &  & 0.0016 & 0.2971 & 1.9343 & 0.6291 & 6.8266 & 0.3463 & 0.6724 & 23.1163 \\
 &  & 0.0022 & 0.3293 & 1.9872 & 0.6679 & 6.3829 & 0.3327 & 0.5117 & 22.9238 \\
\midrule
Contextual & $\eta$ & $10^{7}$ & 0.2697 & 1.8860 & 0.5978 & 7.4913 & 0.3608 & 0.9545 & 23.3394 \\
 &  & $10^{8}$ & 0.3119 & 1.9629 & 0.6863 & 7.5931 & 0.3544 & 0.9486 & 23.2936 \\
 &  & $10^{9}$ & 0.3163 & 1.9603 & 0.6892 & 7.5864 & 0.3536 & 0.9431 & 23.2692 \\
 &  & \textbf{$10^{10}$} & 0.3225 & 1.9793 & 0.6994 & 7.6795 & 0.3533 & 0.9296 & 23.2449 \\
\midrule
Shield & radius & 0.8 & 0.2647 & 1.8812 & 0.5934 & 7.5263 & 0.3622 & 0.9659 & 23.3403 \\
 &  & 0.85 & 0.2682 & 1.8903 & 0.5995 & 7.4945 & 0.3620 & 0.9754 & 23.3309 \\
 &  & 0.9 & 0.2716 & 1.8867 & 0.6071 & 7.4683 & 0.3615 & 0.9733 & 23.3185 \\
 &  & \textbf{0.95} & 0.2813 & 1.9112 & 0.6221 & 7.4127 & 0.3624 & 0.9734 & 23.2872 \\
 &  & 1 & 0.2938 & 1.9402 & 0.6459 & 7.2212 & 0.3629 & 0.9532 & 23.2202 \\
\midrule
CNO & learning rate & \textbf{0.01} & 0.2865 & 1.9288 & 0.6416 & 7.3931 & 0.3642 & 1.0084 & 23.3135 \\
 &  & 0.013 & 0.2993 & 1.9511 & 0.6490 & 7.2446 & 0.3636 & 0.9870 & 23.2248 \\
 &  & 0.017 & 0.3211 & 1.9990 & 0.6735 & 7.0278 & 0.3633 & 0.9264 & 23.1272 \\
 &  & 0.02 & 0.3402 & 2.0102 & 0.6986 & 6.8158 & 0.3627 & 0.8912 & 23.0702 \\
\midrule
Group & $M$ & 16 & 0.2845 & 1.9147 & 0.6214 & 7.4720 & 0.3620 & 0.9337 & 23.3070 \\
 &  & \textbf{32} & 0.3045 & 1.9727 & 0.6325 & 7.4173 & 0.3620 & 0.9264 & 23.2908 \\
 &  & 64 & 0.3346 & 2.0263 & 0.6605 & 7.3126 & 0.3623 & 0.9378 & 23.2889 \\
\bottomrule
\end{tabular}
\end{table}

\begin{table}[t]
\centering
\scriptsize
\setlength{\tabcolsep}{4pt}
\caption{All GenEval sweep results of \cref{fig:geneval_swap_comparison} for FLUX.2-Klein-9B. Bold settings are used in \cref{tab:comparison}.}
\label{tab:geneval_sweep_flux2k}
\begin{tabular}{@{}llr ccccccc@{}}
\toprule
Method & Parameter & Value & DreamSim$\uparrow$ & Vendi$\uparrow$ & LPIPS$\uparrow$ & HPSv3$\uparrow$ & CLIP$\uparrow$ & ImageReward$\uparrow$ & PickScore$\uparrow$ \\
\midrule
Baseline & -- & -- & 0.1964 & 1.6791 & 0.5495 & 9.0595 & 0.3876 & 1.3912 & 23.6764 \\
\midrule
Ours & $\delta$ & 20 & 0.2534 & 1.8085 & 0.6402 & 9.0646 & 0.3871 & 1.3800 & 23.6484 \\
 &  & \textbf{40} & 0.2643 & 1.8280 & 0.6599 & 8.9825 & 0.3860 & 1.3430 & 23.6246 \\
 &  & 60 & 0.2799 & 1.8545 & 0.6833 & 8.7870 & 0.3853 & 1.3175 & 23.5176 \\
 &  & 80 & 0.2806 & 1.8506 & 0.6876 & 8.7634 & 0.3854 & 1.3077 & 23.5003 \\
\midrule
NegToMe & $\alpha$ & 0.5 & 0.2280 & 1.7577 & 0.6224 & 8.9915 & 0.3863 & 1.3694 & 23.6605 \\
 &  & \textbf{0.75} & 0.2426 & 1.7955 & 0.6373 & 8.8926 & 0.3857 & 1.3454 & 23.6210 \\
 &  & 1 & 0.2533 & 1.8083 & 0.6455 & 8.6861 & 0.3847 & 1.3394 & 23.5745 \\
 &  & 1.45 & 0.2732 & 1.8431 & 0.6523 & 8.4658 & 0.3834 & 1.2835 & 23.4784 \\
\midrule
PG & coefficient & 0.2 & 0.1998 & 1.6773 & 0.5689 & 9.0421 & 0.3869 & 1.3937 & 23.6170 \\
 &  & \textbf{0.3} & 0.2031 & 1.6799 & 0.5787 & 8.9348 & 0.3855 & 1.3889 & 23.4989 \\
 &  & 0.4 & 0.2047 & 1.6895 & 0.5893 & 8.6694 & 0.3835 & 1.3731 & 23.3188 \\
 &  & 0.49 & 0.2060 & 1.7031 & 0.5995 & 8.2699 & 0.3816 & 1.3499 & 23.1134 \\
\midrule
CADS & noise scale & 0.012 & 0.2069 & 1.7159 & 0.5604 & 9.0639 & 0.3866 & 1.3997 & 23.6743 \\
 &  & \textbf{0.022} & 0.2258 & 1.7567 & 0.5711 & 8.9555 & 0.3859 & 1.3712 & 23.6665 \\
 &  & 0.032 & 0.2438 & 1.7995 & 0.5846 & 8.6919 & 0.3820 & 1.3413 & 23.5996 \\
 &  & 0.043 & 0.2781 & 1.8667 & 0.6011 & 8.2460 & 0.3751 & 1.2525 & 23.4435 \\
\midrule
Contextual & $\eta$ & $2.5\!\times\!10^{8}$ & 0.2034 & 1.7016 & 0.5601 & 9.0719 & 0.3873 & 1.3979 & 23.6734 \\
 &  & $5\!\times\!10^{8}$ & 0.2236 & 1.7348 & 0.5827 & 9.0345 & 0.3870 & 1.3934 & 23.6843 \\
 &  & \textbf{$7.5\!\times\!10^{8}$} & 0.2519 & 1.8037 & 0.6093 & 8.9387 & 0.3859 & 1.3736 & 23.6401 \\
 &  & $10^{9}$ & 0.2990 & 1.8880 & 0.6404 & 8.6567 & 0.3823 & 1.3217 & 23.5370 \\
\midrule
Shield & radius & 1.1 & 0.2110 & 1.7054 & 0.5911 & 9.1145 & 0.3876 & 1.3960 & 23.6494 \\
 &  & 1.27 & 0.2262 & 1.7242 & 0.6387 & 9.0618 & 0.3887 & 1.3885 & 23.5361 \\
 &  & \textbf{1.44} & 0.2386 & 1.7401 & 0.6735 & 8.8634 & 0.3887 & 1.3902 & 23.3728 \\
 &  & 1.61 & 0.2489 & 1.7499 & 0.7029 & 8.5748 & 0.3876 & 1.3686 & 23.1534 \\
\midrule
CNO & learning rate & 0.05 & 0.2021 & 1.7008 & 0.5799 & 9.0836 & 0.3884 & 1.3972 & 23.6817 \\
 &  & 0.09 & 0.2047 & 1.7054 & 0.5935 & 8.9769 & 0.3885 & 1.3926 & 23.6116 \\
 &  & \textbf{0.13} & 0.2070 & 1.7079 & 0.6032 & 8.8135 & 0.3891 & 1.3840 & 23.5130 \\
 &  & 0.18 & 0.2090 & 1.7149 & 0.6161 & 8.4981 & 0.3887 & 1.3525 & 23.3423 \\
\midrule
Group & $M$ & 8 & 0.1949 & 1.6704 & 0.5487 & 9.0846 & 0.3879 & 1.3994 & 23.6893 \\
 &  & 16 & 0.2132 & 1.7236 & 0.5792 & 8.9515 & 0.3861 & 1.3934 & 23.6194 \\
 &  & 32 & 0.2225 & 1.7427 & 0.6007 & 8.9414 & 0.3867 & 1.3859 & 23.6697 \\
 &  & \textbf{64} & 0.2410 & 1.7915 & 0.6157 & 8.8540 & 0.3861 & 1.3517 & 23.6074 \\
\bottomrule
\end{tabular}
\end{table}

\begin{table}[t]
\centering
\scriptsize
\setlength{\tabcolsep}{4pt}
\caption{All GenEval sweep results of \cref{fig:geneval_swap_comparison} for SDXL. Bold settings are used in \cref{tab:comparison}.}
\label{tab:geneval_sweep_sdxl}
\begin{tabular}{@{}llr ccccccc@{}}
\toprule
Method & Parameter & Value & DreamSim$\uparrow$ & Vendi$\uparrow$ & LPIPS$\uparrow$ & HPSv3$\uparrow$ & CLIP$\uparrow$ & ImageReward$\uparrow$ & PickScore$\uparrow$ \\
\midrule
Baseline & -- & -- & 0.3212 & 1.9565 & 0.7243 & 5.9234 & 0.3828 & 0.9481 & 22.9723 \\
\midrule
Ours & $\delta$ & 5 & 0.3459 & 2.0288 & 0.7374 & 5.8322 & 0.3829 & 0.9508 & 22.9027 \\
 &  & \textbf{10} & 0.3700 & 2.0863 & 0.7480 & 5.6233 & 0.3804 & 0.8938 & 22.7996 \\
 &  & 15 & 0.3851 & 2.1091 & 0.7545 & 5.4437 & 0.3780 & 0.8587 & 22.7336 \\
 &  & 20 & 0.3892 & 2.1318 & 0.7575 & 5.4396 & 0.3796 & 0.8619 & 22.7217 \\
\midrule
NegToMe & $\alpha$ & 0.07 & 0.3341 & 1.9886 & 0.7360 & 5.9201 & 0.3827 & 0.9410 & 22.9368 \\
 &  & \textbf{0.14} & 0.3393 & 2.0042 & 0.7459 & 5.7745 & 0.3822 & 0.9311 & 22.8969 \\
 &  & 0.21 & 0.3591 & 2.0546 & 0.7599 & 5.4890 & 0.3798 & 0.8662 & 22.7609 \\
 &  & 0.28 & 0.3830 & 2.1180 & 0.7678 & 4.8133 & 0.3761 & 0.7780 & 22.5288 \\
\midrule
SPARKE & guidance scale & 0.034 & 0.3280 & 1.9691 & 0.7408 & 5.9558 & 0.3823 & 0.9527 & 22.9608 \\
 &  & 0.069 & 0.3336 & 1.9764 & 0.7560 & 5.8783 & 0.3810 & 0.9429 & 22.9329 \\
 &  & \textbf{0.1} & 0.3388 & 1.9795 & 0.7713 & 5.7089 & 0.3797 & 0.9398 & 22.8326 \\
 &  & 0.14 & 0.3433 & 1.9888 & 0.7834 & 4.9676 & 0.3757 & 0.8894 & 22.5126 \\
\midrule
PG & coefficient & 0.001 & 0.3243 & 1.9599 & 0.7349 & 5.9434 & 0.3822 & 0.9561 & 22.9302 \\
 &  & 0.002 & 0.3258 & 1.9532 & 0.7453 & 5.8878 & 0.3826 & 0.9742 & 22.8620 \\
 &  & \textbf{0.003} & 0.3299 & 1.9598 & 0.7531 & 5.6796 & 0.3811 & 0.9589 & 22.7121 \\
 &  & 0.004 & 0.3321 & 1.9654 & 0.7614 & 5.2724 & 0.3799 & 0.9623 & 22.5356 \\
\midrule
CADS & noise scale & 0.005 & 0.3329 & 1.9849 & 0.7258 & 5.9188 & 0.3831 & 0.9394 & 22.9618 \\
 &  & 0.01 & 0.3433 & 2.0016 & 0.7249 & 5.8180 & 0.3811 & 0.8927 & 22.9438 \\
 &  & \textbf{0.015} & 0.3595 & 2.0310 & 0.7264 & 5.5772 & 0.3786 & 0.8433 & 22.8572 \\
 &  & 0.02 & 0.3693 & 2.0729 & 0.7342 & 5.4113 & 0.3790 & 0.8011 & 22.7876 \\
\midrule
Shield & radius & 0.9 & 0.3239 & 1.9728 & 0.7298 & 5.9446 & 0.3826 & 0.9508 & 22.9591 \\
 &  & 0.95 & 0.3282 & 1.9791 & 0.7348 & 5.8620 & 0.3821 & 0.9353 & 22.9455 \\
 &  & 1 & 0.3290 & 1.9795 & 0.7420 & 5.7864 & 0.3822 & 0.9351 & 22.9116 \\
 &  & \textbf{1.05} & 0.3343 & 1.9852 & 0.7604 & 5.5742 & 0.3819 & 0.9180 & 22.7516 \\
\midrule
CNO & learning rate & 0.17 & 0.3240 & 1.9612 & 0.7351 & 5.8812 & 0.3827 & 0.9368 & 22.9732 \\
 &  & 0.34 & 0.3338 & 1.9854 & 0.7444 & 5.8319 & 0.3825 & 0.9349 & 22.9413 \\
 &  & \textbf{0.51} & 0.3542 & 2.0364 & 0.7587 & 5.5920 & 0.3820 & 0.9154 & 22.8427 \\
 &  & 0.68 & 0.3750 & 2.0484 & 0.7770 & 5.2773 & 0.3801 & 0.8631 & 22.7012 \\
\midrule
Group & $M$ & 8 & 0.3506 & 2.0503 & 0.7404 & 5.9582 & 0.3867 & 1.0012 & 23.0036 \\
 &  & 16 & 0.3779 & 2.0909 & 0.7512 & 5.7041 & 0.3859 & 0.9731 & 22.9213 \\
 &  & \textbf{32} & 0.3969 & 2.1468 & 0.7583 & 5.5191 & 0.3838 & 0.9303 & 22.8413 \\
 &  & 64 & 0.4088 & 2.1757 & 0.7628 & 5.4680 & 0.3850 & 0.8831 & 22.7792 \\
\bottomrule
\end{tabular}
\end{table}

\subsection{Number of Corrected Layers}
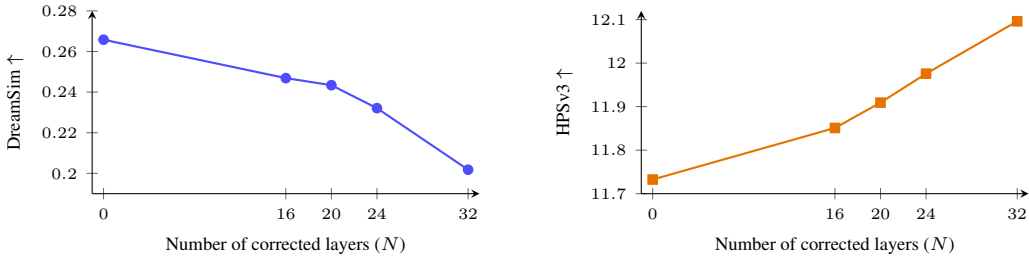
\begin{figure}[h]
    \centering
    \begin{minipage}[t]{0.48\textwidth}
        \centering
        \begin{tikzpicture}[font=\fontfamily{phv}\selectfont]
        \begin{axis}[
            width=\linewidth, height=4.0cm,
            axis lines=left, scaled ticks=false,
            label style={font=\scriptsize}, tick label style={font=\tiny},
            tick style={black!60},
            xmin=-1, xmax=33, ymin=0.19, ymax=0.28,
            xtick={0,16,20,24,32},
            ytick={0.20,0.22,0.24,0.26,0.28},
            yticklabel style={/pgf/number format/fixed, /pgf/number format/precision=2},
            xlabel={Number of corrected layers ($N$)},
            ylabel={DreamSim $\uparrow$},
        ]
        \addplot[myblue, mark=*, mark size=1.6pt, line width=0.8pt] coordinates {
            (0,0.2658) (16,0.2469) (20,0.2434) (24,0.2321) (32,0.2018)
        };
        \end{axis}
        \end{tikzpicture}
    \end{minipage}
    \hfill
    \begin{minipage}[t]{0.48\textwidth}
        \centering
        \begin{tikzpicture}[font=\fontfamily{phv}\selectfont]
        \begin{axis}[
            width=\linewidth, height=4.0cm,
            axis lines=left, scaled ticks=false,
            label style={font=\scriptsize}, tick label style={font=\tiny},
            tick style={black!60},
            xmin=-1, xmax=33, ymin=11.70, ymax=12.12,
            xtick={0,16,20,24,32},
            ytick={11.7,11.8,11.9,12.0,12.1},
            yticklabel style={/pgf/number format/fixed, /pgf/number format/precision=1},
            xlabel={Number of corrected layers ($N$)},
            ylabel={HPSv3 $\uparrow$},
        ]
        \addplot[myorange, mark=square*, mark size=1.6pt, line width=0.8pt] coordinates {
            (0,11.7325) (16,11.8510) (20,11.9092) (24,11.9757) (32,12.0961)
        };
        \end{axis}
        \end{tikzpicture}
    \end{minipage}
    \caption{Effect of correcting the first $N$ transformer layers on DreamSim (left) and HPSv3 (right) for FLUX.2-Klein-9B on DrawBench with $\delta=40^\circ$ and $\bar{\theta}=60^\circ$. TokenRepel is applied at the first sampling step; $N=0$ uses no correction.}
    \label{fig:first_n_blocks_correction}
\end{figure}

As shown in \cref{fig:first_n_blocks_correction}, correcting more layers increases HPSv3 while decreasing DreamSim.
Reducing the number of corrected layers can also reduce the inference computation cost overhead.

\subsection{Alignment of Repulsion and Density Directions}
\begin{figure}[h]
    \centering
    \begin{minipage}[t]{0.48\textwidth}
        \centering
        \begin{tikzpicture}[font=\fontfamily{phv}\selectfont]
        \begin{axis}[
            width=\linewidth, height=4.0cm,
            axis lines=left, scaled ticks=false,
            label style={font=\scriptsize}, tick label style={font=\tiny},
            tick style={black!60},
            xmin=0, xmax=1.8, ymin=-0.33, ymax=0,
            xtick={0,0.4,0.8,1.2,1.6},
            ytick={-0.3,-0.2,-0.1,0},
            yticklabel style={/pgf/number format/fixed, /pgf/number format/precision=1},
            xlabel={Repulsion strength $\tan(\delta)$},
            ylabel={Mean cosine},
        ]
        \addplot[mygreen, mark=*, mark size=1.4pt, line width=0.8pt] coordinates {
            (0.0875,-0.0623) (0.1763,-0.1240) (0.2679,-0.1697) (0.3640,-0.2027)
            (0.4663,-0.2278) (0.5774,-0.2489) (0.7002,-0.2675) (0.8391,-0.2839)
            (1.0000,-0.2944) (1.1918,-0.2984) (1.4281,-0.2998) (1.7321,-0.3014)
        };
        \end{axis}
        \end{tikzpicture}
    \end{minipage}
    \hfill
    \begin{minipage}[t]{0.48\textwidth}
        \centering
        \begin{tikzpicture}[font=\fontfamily{phv}\selectfont]
        \begin{axis}[
            width=\linewidth, height=4.0cm,
            axis lines=left, scaled ticks=false,
            label style={font=\scriptsize}, tick label style={font=\tiny},
            tick style={black!60},
            xmin=0, xmax=1.8, ymin=0.60, ymax=1.0,
            xtick={0,0.4,0.8,1.2,1.6},
            ytick={0.6,0.7,0.8,0.9,1.0},
            yticklabel style={/pgf/number format/fixed, /pgf/number format/precision=1},
            xlabel={Repulsion strength $\tan(\delta)$},
            ylabel={Fraction with cosine $<0$},
        ]
        \addplot[mypurple, mark=square*, mark size=1.4pt, line width=0.8pt] coordinates {
            (0.0875,0.667) (0.1763,0.779) (0.2679,0.847) (0.3640,0.888)
            (0.4663,0.913) (0.5774,0.927) (0.7002,0.939) (0.8391,0.946)
            (1.0000,0.949) (1.1918,0.949) (1.4281,0.949) (1.7321,0.949)
        };
        \end{axis}
        \end{tikzpicture}
    \end{minipage}
    \caption{Alignment between the repulsion displacement and the high-density direction before correction for FLUX.2-Klein-9B on DrawBench with $\bar{\theta}=60^\circ$. The plots show the mean token-wise cosine (left) and the fraction of negatively aligned tokens (right) as $\delta$ varies.}
    \label{fig:token_density_alignment_sigma}
\end{figure}
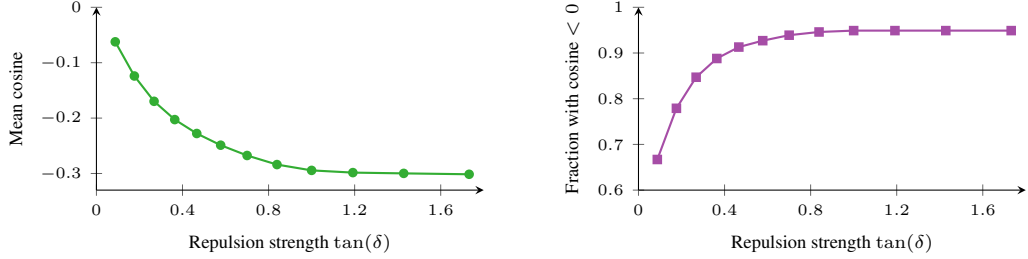

To examine when token density preservation acts,
we measure the cosine similarity between the repulsion displacement $\ubf_m^r-\ubf_m^+$ and the high-density direction $\ubf_m^+-\ubf_m^-$ before correction.
We average over tokens with nonzero repulsion displacement across the layers at the first sampling step of FLUX.2-Klein-9B,
including text tokens in single-stream layers.
As $\delta$ increases, the mean cosine becomes more negative and the fraction of negatively aligned tokens rises (\cref{fig:token_density_alignment_sigma}).
These tokens oppose the estimated high-density direction and would be modified by the half-space correction in \cref{eq:repel_plus_correct}.

\subsection{Feature Visualization of Token Density Preservation}
\Cref{fig:visual_patch_diff_p060} shows one example of how token density preservation affects the generated images and the intermediate features on FLUX.2-Klein-9B.
\input{figures_arxiv/analysis_ablation_plots/correction_visual_draw_flux2k_p060}

\subsection{More Qualitative Results}
\begin{figure}[p]
\centering
\begin{tikzpicture}[font=\fontfamily{phv}\selectfont, inner sep=0pt, outer sep=0pt]
    \def\nrows{4}
    \pgfmathsetlengthmacro{\figw}{\linewidth}
    \pgfmathsetlengthmacro{\imgsep}{1pt}
    \pgfmathsetlengthmacro{\groupsep}{4pt}
    \pgfmathsetlengthmacro{\pad}{3pt}
    \pgfmathsetlengthmacro{\imgw}{(\figw - 6*\imgsep - 4*\pad - \groupsep)/8}
    \pgfmathsetlengthmacro{\groupw}{4*\imgw + 3*\imgsep}
    \pgfmathsetlengthmacro{\xbase}{\pad}
    \pgfmathsetlengthmacro{\xours}{\groupw + 3*\pad + \groupsep}
    \pgfmathsetlengthmacro{\rowh}{\imgw + 15pt}

    \pgfmathsetlengthmacro{\boxtop}{16pt}
    \pgfmathsetlengthmacro{\boxbot}{-(\nrows - 1)*\rowh - \imgw - 16pt}
    \fill[blue!8, rounded corners=4pt] (0, \boxtop) rectangle ({\groupw + 2*\pad}, \boxbot);
    \fill[yellow!25, rounded corners=4pt] ({\xours - \pad}, \boxtop) rectangle (\figw, \boxbot);

    \newcommand{\qualrow}[4]{
        \pgfmathsetlengthmacro{\y}{-#1*\rowh}
        \foreach \i in {0,...,3} {
            \node[anchor=north west] at ({\xbase + \i*(\imgw + \imgsep)}, \y)
                {\includegraphics[width=\imgw]{figures_arxiv/qualitative_show/images/#2_\i.jpg}};
            \node[anchor=north west] at ({\xours + \i*(\imgw + \imgsep)}, \y)
                {\includegraphics[width=\imgw]{figures_arxiv/qualitative_show/images/#3_\i.jpg}};
        }
        \node[anchor=north, fill=white, rounded corners=2pt, inner xsep=4pt, inner ysep=1.5pt,
              font=\fontfamily{phv}\selectfont\scriptsize]
            at ({\figw/2}, {\y - \imgw - 3pt}) {\textit{``#4''}};
    }

    \node[anchor=south, font=\fontfamily{phv}\selectfont\small]
        at ({\xbase + \groupw/2}, 3pt) {\textbf{Baseline}};
    \node[anchor=south, font=\fontfamily{phv}\selectfont\small]
        at ({\xours + \groupw/2}, 3pt) {\textbf{With Ours}};

    \qualrow{0}{flux2k_base_p019}{flux2k_ours_p019}{An owl wearing round glasses reading a newspaper.}
    \qualrow{1}{flux2k_base_p009}{flux2k_ours_p009}{A teddy bear astronaut floating in space.}
    \qualrow{2}{flux2k_base_p033}{flux2k_ours_p033}{A grandmother knitting by the fireplace with a cat.}
    \qualrow{3}{flux2k_base_p024}{flux2k_ours_p024}{A giant whale swimming above a city.}
\end{tikzpicture}
\caption{FLUX.2-Klein-9B qualitative results on additional prompts with $\delta=80^\circ$, under the same initial noise.}
\label{fig:qual_show_flux2k}
\end{figure}
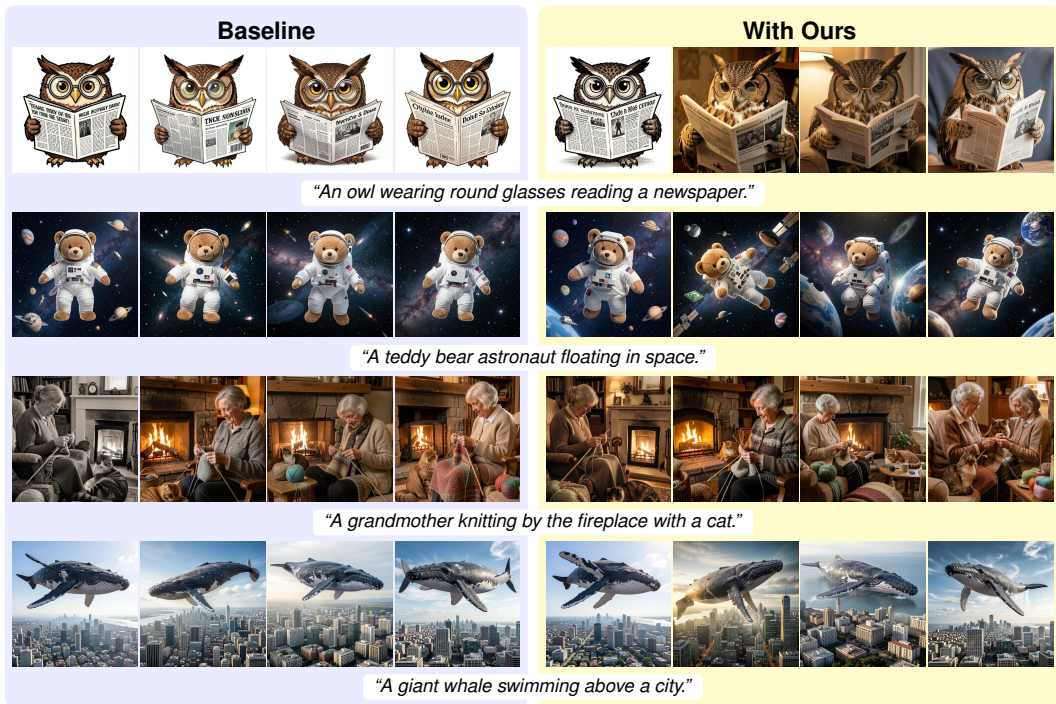

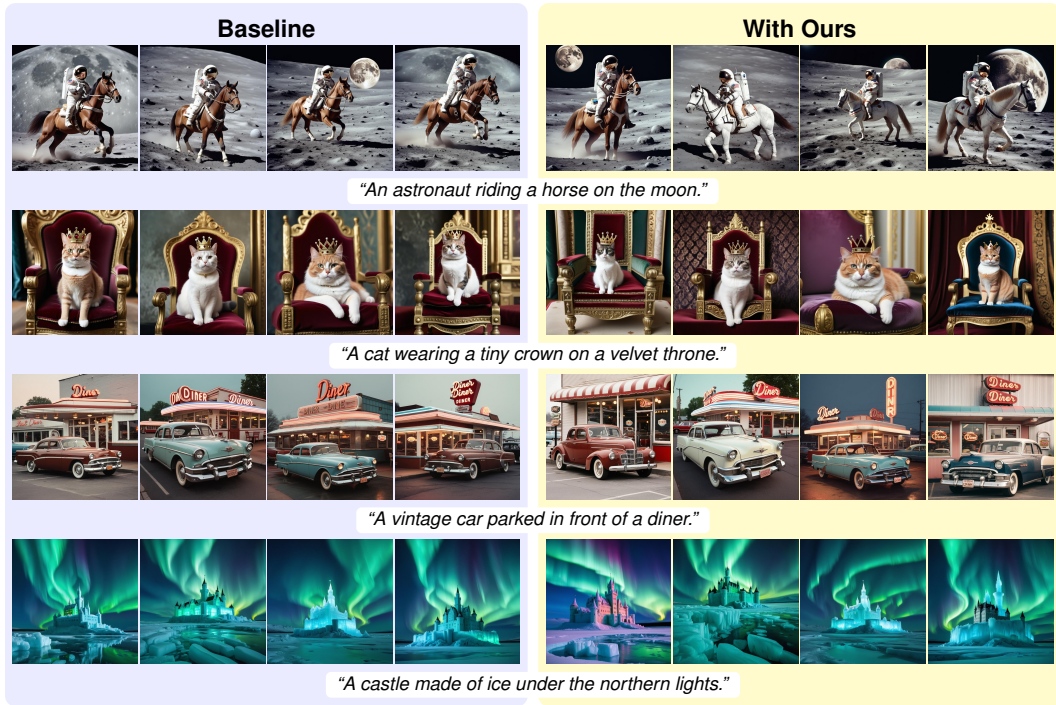
\begin{figure}[p]
\centering
\begin{tikzpicture}[font=\fontfamily{phv}\selectfont, inner sep=0pt, outer sep=0pt]
    \def\nrows{4}
    \pgfmathsetlengthmacro{\figw}{\linewidth}
    \pgfmathsetlengthmacro{\imgsep}{1pt}
    \pgfmathsetlengthmacro{\groupsep}{4pt}
    \pgfmathsetlengthmacro{\pad}{3pt}
    \pgfmathsetlengthmacro{\imgw}{(\figw - 6*\imgsep - 4*\pad - \groupsep)/8}
    \pgfmathsetlengthmacro{\groupw}{4*\imgw + 3*\imgsep}
    \pgfmathsetlengthmacro{\xbase}{\pad}
    \pgfmathsetlengthmacro{\xours}{\groupw + 3*\pad + \groupsep}
    \pgfmathsetlengthmacro{\rowh}{\imgw + 15pt}

    \pgfmathsetlengthmacro{\boxtop}{16pt}
    \pgfmathsetlengthmacro{\boxbot}{-(\nrows - 1)*\rowh - \imgw - 16pt}
    \fill[blue!8, rounded corners=4pt] (0, \boxtop) rectangle ({\groupw + 2*\pad}, \boxbot);
    \fill[yellow!25, rounded corners=4pt] ({\xours - \pad}, \boxtop) rectangle (\figw, \boxbot);

    \newcommand{\qualrow}[4]{
        \pgfmathsetlengthmacro{\y}{-#1*\rowh}
        \foreach \i in {0,...,3} {
            \node[anchor=north west] at ({\xbase + \i*(\imgw + \imgsep)}, \y)
                {\includegraphics[width=\imgw]{figures_arxiv/qualitative_show/images/#2_\i.jpg}};
            \node[anchor=north west] at ({\xours + \i*(\imgw + \imgsep)}, \y)
                {\includegraphics[width=\imgw]{figures_arxiv/qualitative_show/images/#3_\i.jpg}};
        }
        \node[anchor=north, fill=white, rounded corners=2pt, inner xsep=4pt, inner ysep=1.5pt,
              font=\fontfamily{phv}\selectfont\scriptsize]
            at ({\figw/2}, {\y - \imgw - 3pt}) {\textit{``#4''}};
    }

    \node[anchor=south, font=\fontfamily{phv}\selectfont\small]
        at ({\xbase + \groupw/2}, 3pt) {\textbf{Baseline}};
    \node[anchor=south, font=\fontfamily{phv}\selectfont\small]
        at ({\xours + \groupw/2}, 3pt) {\textbf{With Ours}};

    \qualrow{0}{sdxl_base_p030}{sdxl_ours_p030}{An astronaut riding a horse on the moon.}
    \qualrow{1}{sdxl_base_p002}{sdxl_ours_p002}{A cat wearing a tiny crown on a velvet throne.}
    \qualrow{2}{sdxl_base_p012}{sdxl_ours_p012}{A vintage car parked in front of a diner.}
    \qualrow{3}{sdxl_base_p021}{sdxl_ours_p021}{A castle made of ice under the northern lights.}
\end{tikzpicture}
\caption{SDXL qualitative results on additional prompts with $\delta=25^\circ$, under the same initial noise.}
\label{fig:qual_show_sdxl}
\end{figure}

\begin{figure}[p]
\centering
\begin{tikzpicture}[font=\fontfamily{phv}\selectfont, inner sep=0pt, outer sep=0pt]
    \pgfmathsetlengthmacro{\figw}{\linewidth}
    \pgfmathsetlengthmacro{\imgsep}{1pt}
    \pgfmathsetlengthmacro{\groupsep}{6pt}
    \pgfmathsetlengthmacro{\pad}{3pt}
    \pgfmathsetlengthmacro{\groupw}{(\figw - \groupsep)/2}
    \pgfmathsetlengthmacro{\imgw}{(\groupw - 2*\pad - 3*\imgsep)/4}
    \pgfmathsetlengthmacro{\labelh}{10pt}
    \pgfmathsetlengthmacro{\rowh}{\imgw + \labelh + 4pt}
    \pgfmathsetlengthmacro{\boxbot}{-6*\rowh - 16pt}

    \newcommand{\deltagroup}[3]{
        \fill[green!10, rounded corners=4pt] (#1, 0) rectangle ({#1 + \groupw}, \boxbot);
        \foreach \d [count=\r from 0] in {0,20,30,40,50,60} {
            \pgfmathsetlengthmacro{\y}{-\r*\rowh - \pad}
            \node[anchor=north, font=\fontfamily{phv}\selectfont\scriptsize]
                at ({#1 + \groupw/2}, \y) {$\delta=\d^\circ$\ifnum\d=0\ (Baseline)\fi};
            \foreach \i in {0,...,3} {
                \node[anchor=north west] at ({#1 + \pad + \i*(\imgw + \imgsep)}, {\y - \labelh})
                    {\includegraphics[width=\imgw]{figures_arxiv/qualitative_delta/images/#2_d\d_\i.jpg}};
            }
        }
        \node[anchor=north, fill=white, rounded corners=2pt, inner xsep=4pt, inner ysep=1.5pt,
              font=\fontfamily{phv}\selectfont\scriptsize]
            at ({#1 + \groupw/2}, {-6*\rowh - 2pt}) {\textit{``#3''}};
    }

    \deltagroup{0pt}{flux_p014}{A ballerina dancing in an empty train station.}
    \deltagroup{\groupw + \groupsep}{flux_p037}{A glass bottle with a tiny galaxy inside.}
\end{tikzpicture}
\caption{FLUX.1-dev results as the token rotation angle $\delta$ increases, under the same initial noise. $\delta=0^\circ$ is the baseline.
Diversity increases with $\delta$.}
\label{fig:qual_delta_flux}
\end{figure}
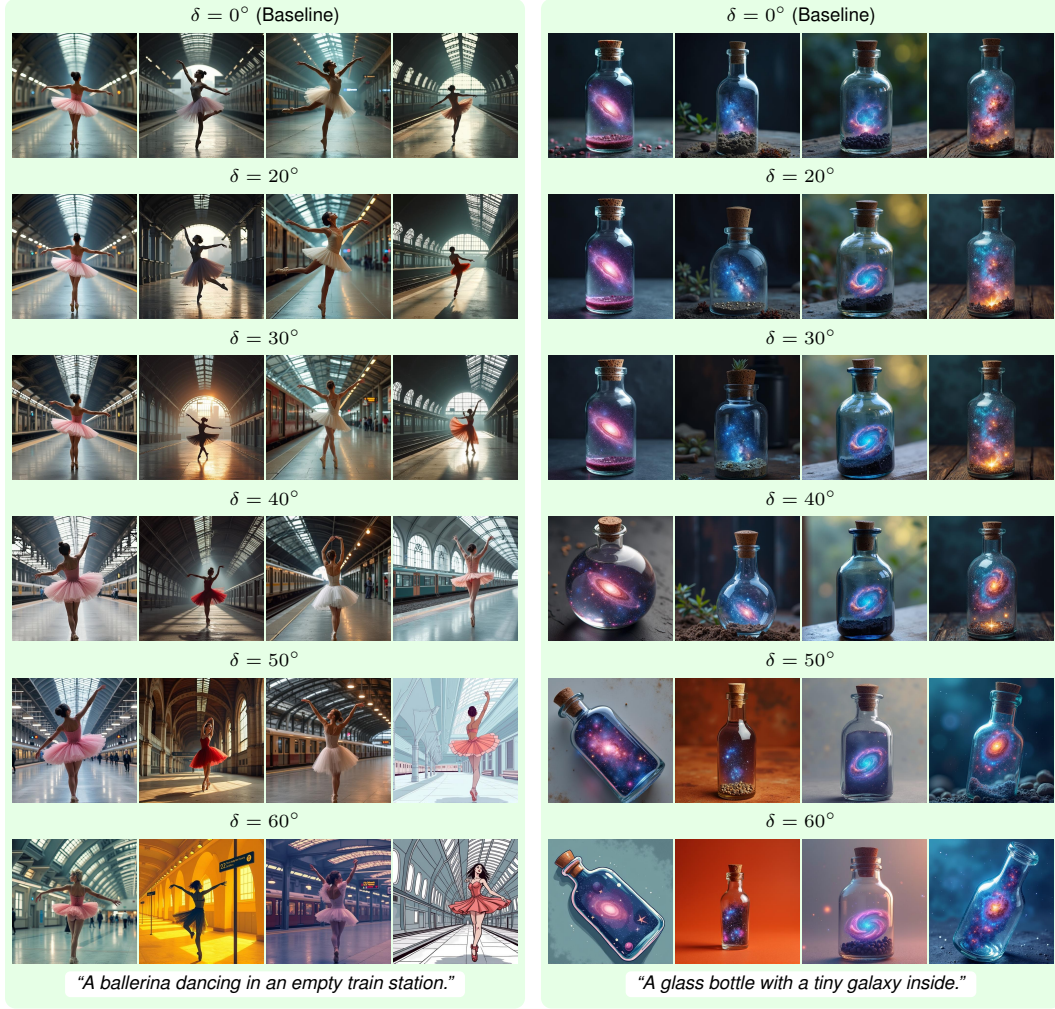

\Cref{fig:qual_show_flux,fig:qual_show_flux2k,fig:qual_show_sdxl} show results on additional prompts beyond GenEval. 
\Cref{fig:qual_geneval_app_flux,fig:qual_geneval_app_flux2k,fig:qual_geneval_app_sdxl} present more GenEval examples of the baseline and TokenRepel at the selected settings in \cref{tab:comparison}.
\Cref{fig:kontext_qua_appendix,fig:kontext_qua_appendix2} show additional image-editing results on KontextBench using FLUX Kontext.
\Cref{fig:qual_delta_flux} shows how the generated images change as $\delta$ increases from $0^\circ$ to $60^\circ$ on FLUX.1-dev.
\begin{figure}[p]
\centering
\begin{tikzpicture}[font=\fontfamily{phv}\selectfont, inner sep=0pt, outer sep=0pt]
    \def\nrows{4}
    \pgfmathsetlengthmacro{\figw}{\linewidth}
    \pgfmathsetlengthmacro{\imgsep}{1pt}
    \pgfmathsetlengthmacro{\groupsep}{4pt}
    \pgfmathsetlengthmacro{\pad}{3pt}
    \pgfmathsetlengthmacro{\imgw}{(\figw - 6*\imgsep - 4*\pad - \groupsep)/8}
    \pgfmathsetlengthmacro{\groupw}{4*\imgw + 3*\imgsep}
    \pgfmathsetlengthmacro{\xbase}{\pad}
    \pgfmathsetlengthmacro{\xours}{\groupw + 3*\pad + \groupsep}
    \pgfmathsetlengthmacro{\rowh}{\imgw + 15pt}

    \pgfmathsetlengthmacro{\boxtop}{16pt}
    \pgfmathsetlengthmacro{\boxbot}{-(\nrows - 1)*\rowh - \imgw - 16pt}
    \fill[blue!8, rounded corners=4pt] (0, \boxtop) rectangle ({\groupw + 2*\pad}, \boxbot);
    \fill[yellow!25, rounded corners=4pt] ({\xours - \pad}, \boxtop) rectangle (\figw, \boxbot);

    \newcommand{\qualrow}[4]{
        \pgfmathsetlengthmacro{\y}{-#1*\rowh}
        \foreach \i in {0,...,3} {
            \node[anchor=north west] at ({\xbase + \i*(\imgw + \imgsep)}, \y)
                {\includegraphics[width=\imgw]{figures_arxiv/qualitative_show/images/#2_\i.jpg}};
            \node[anchor=north west] at ({\xours + \i*(\imgw + \imgsep)}, \y)
                {\includegraphics[width=\imgw]{figures_arxiv/qualitative_show/images/#3_\i.jpg}};
        }
        \node[anchor=north, fill=white, rounded corners=2pt, inner xsep=4pt, inner ysep=1.5pt,
              font=\fontfamily{phv}\selectfont\scriptsize]
            at ({\figw/2}, {\y - \imgw - 3pt}) {\textit{``#4''}};
    }

    \node[anchor=south, font=\fontfamily{phv}\selectfont\small]
        at ({\xbase + \groupw/2}, 3pt) {\textbf{Baseline}};
    \node[anchor=south, font=\fontfamily{phv}\selectfont\small]
        at ({\xours + \groupw/2}, 3pt) {\textbf{With Ours}};

    \qualrow{0}{flux_base_p008}{flux_ours_p008}{A samurai standing in a field of red flowers.}
    \qualrow{1}{flux_base_p010}{flux_ours_p010}{A wizard reading a glowing book in a library.}
    \qualrow{2}{flux_base_p037}{flux_ours_p037}{A glass bottle with a tiny galaxy inside.}
    \qualrow{3}{flux_base_p041}{flux_ours_p041}{A jazz musician playing saxophone under a streetlight.}
\end{tikzpicture}
\caption{FLUX.1-dev qualitative results on additional prompts with $\delta=50^\circ$, under the same initial noise.}
\label{fig:qual_show_flux}
\end{figure}
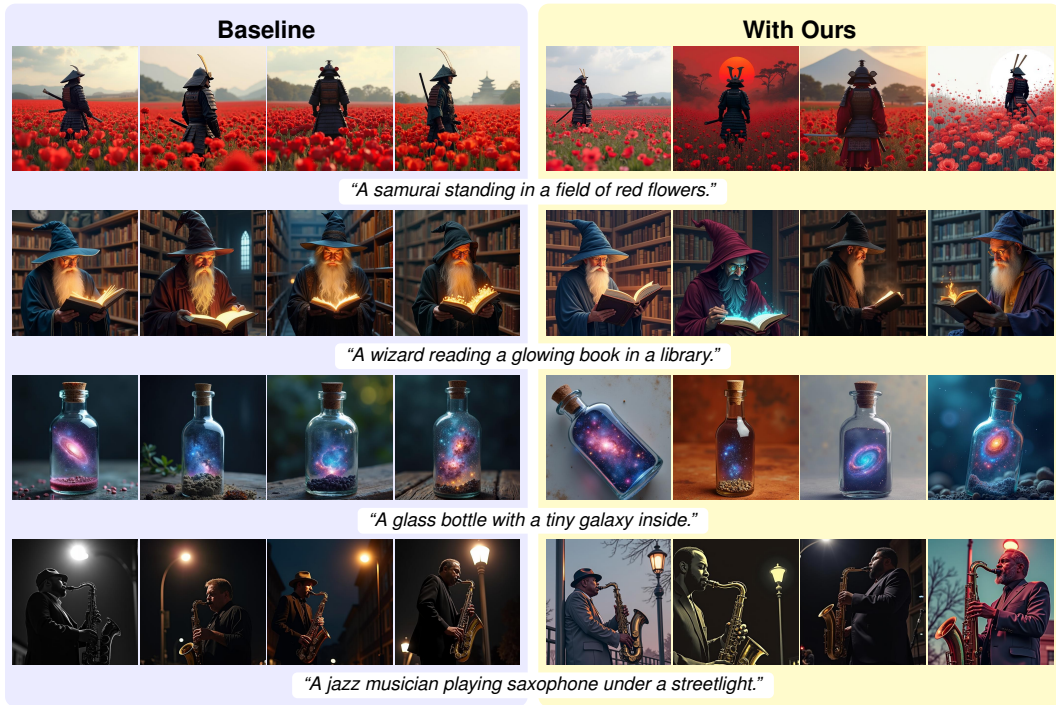

\input{figures_arxiv/qualitative_geneval/qualitative_geneval_appendix_flux}
\input{figures_arxiv/qualitative_geneval/qualitative_geneval_appendix_flux2k}
\input{figures_arxiv/qualitative_geneval/qualitative_geneval_appendix_sdxl}
\newcommand{\kontextrow}[4]{%
\makebox[1.1em]{\rotatebox{90}{\makebox[#4]{\scriptsize #2}}}%
\includegraphics[#3]{assets/kontext_appendix/#1_img0.jpg}\hspace{1pt}%
\includegraphics[#3]{assets/kontext_appendix/#1_img1.jpg}\hspace{1pt}%
\includegraphics[#3]{assets/kontext_appendix/#1_img2.jpg}\hspace{1pt}%
\includegraphics[#3]{assets/kontext_appendix/#1_img3.jpg}}
\newlength{\kontextw}
\newcommand{\kontextcell}[4]{%
\begin{minipage}{0.8\textwidth}
\setlength{\kontextw}{\linewidth}%
\raggedright\fontfamily{phv}\selectfont
\hspace{1.1em}%
\begin{tabular}{@{}c@{}}
\includegraphics[#3]{assets/kontext_appendix/#1/source.jpg}\\[-1pt]
{\scriptsize Source}
\end{tabular}%
\hspace{8pt}%
\begin{minipage}[c]{0.5\kontextw}
\small ``#2''
\end{minipage}\\[5pt]
\kontextrow{#1/baseline}{Baseline}{#3}{#4}\\[2pt]
\kontextrow{#1/ours}{Ours}{#3}{#4}
\end{minipage}}

\begin{figure}[t]
\centering
\kontextcell{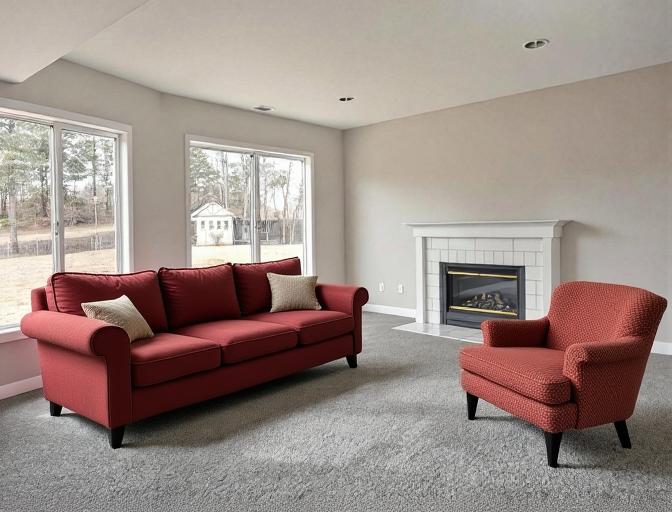}{add ugly awful furniture to the room}{width=0.232\kontextw}{1.8cm}\\[16pt]
\kontextcell{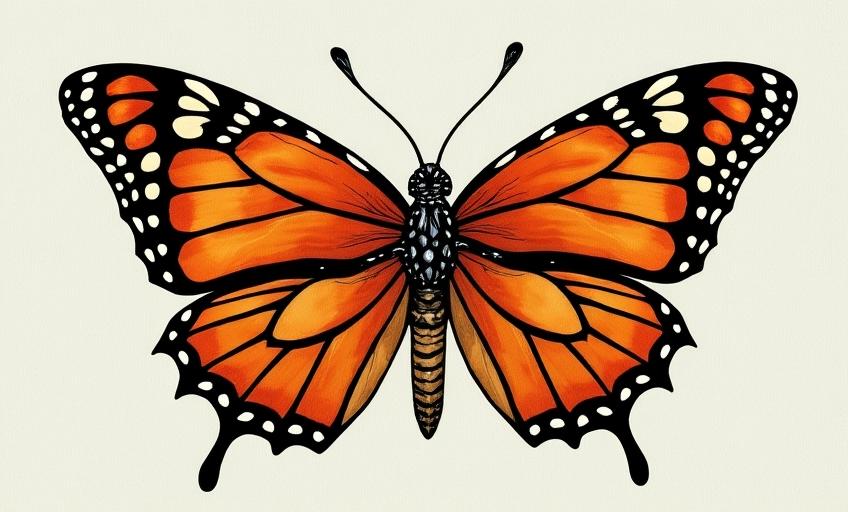}{art in this style of a butterfly}{width=0.232\kontextw}{1.45cm}
\caption{Qualitative image-editing results on KontextBench using FLUX Kontext under the same initial noise.
Each example shows the source image and the instruction (top),
followed by four edits from the baseline and TokenRepel.}
\label{fig:kontext_qua_appendix}
\end{figure}

\begin{figure}[t]
\centering
\kontextcell{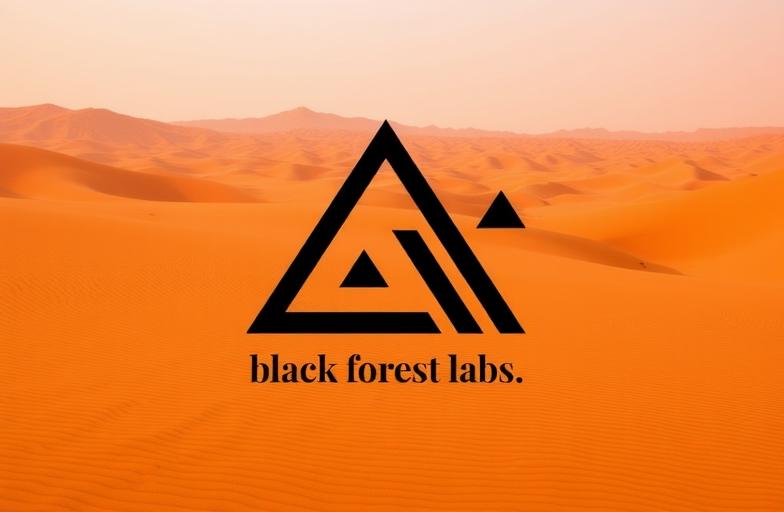}{Replace the background with a desert}{width=0.232\kontextw}{1.55cm}\\[16pt]
\kontextcell{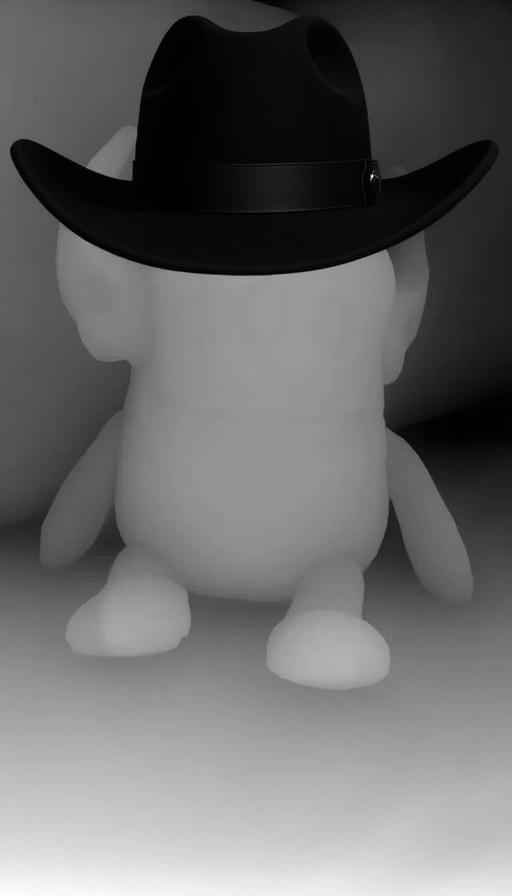}{add a cowboy hat}{height=2.6cm}{2.6cm}
\caption{Qualitative image-editing results on KontextBench using FLUX Kontext under the same initial noise.
Each example shows the source image and the instruction (top),
followed by four edits from the baseline and TokenRepel.}
\label{fig:kontext_qua_appendix2}
\end{figure}

\section{Derivation of Token Density Preservation}
\label{app:token_density_constraint}
We seek the smallest change to the repulsion-induced velocity difference $\Delta^r_{ij}$ that places it in the half-space $\mathcal{R}_{ij}$:
\begin{align}
    \min_{\bar{\Delta}^r_{ij}} \quad
    \frac{1}{2}\|\bar{\Delta}^r_{ij}-\Delta^r_{ij}\|^2
    \quad \text{s.t.} \quad
    \langle \bar{\Delta}^r_{ij},\Delta^h_{ij}\rangle \geq 0.
    \label{eq:token_density_constraint_problem}
\end{align}
If $\langle\Delta^r_{ij},\Delta^h_{ij}\rangle\geq 0$, the original update is feasible and already minimizes the objective.
Otherwise, the optimum lies on the boundary $\langle\bar{\Delta}^r_{ij},\Delta^h_{ij}\rangle=0$.
The stationarity condition gives $\bar{\Delta}^r_{ij}=\Delta^r_{ij}+\lambda\Delta^h_{ij}$, where
$\lambda=-\langle\Delta^r_{ij},\Delta^h_{ij}\rangle/\|\Delta^h_{ij}\|^2$.
Substituting $\lambda$ yields the projection in \cref{eq:repel_plus_correct}.

\section{Additional Implementation Details}
\label{app:implementation_details}

\subsection{Parameter Settings}
\label{app:parameter_settings}

\paragraph{Common setup.}
We generate 4 images per prompt at $512\times512$ with seed 0 and share the initial noise across methods; images are saved as JPEG (quality 75) before evaluation.
FLUX.1-dev uses guidance 3.5, and SDXL uses CFG 7.5.
The SDXL checkpoint is RunDiffusion/Juggernaut-XI-v11.
The negative prompt used for all models is:
\textit{``blurry, low quality, unrealistic, distorted, deformed, bad anatomy, ugly, watermark, oversaturated, flat background, simple environment, cartoon, anime, low detail, disfigured, disgusting, noise, overexposed, underexposed, compressed artifacts, plastic skin, airbrushed, doll-like, fake, low resolution, illustration, out of focus.''}

\paragraph{TokenRepel.}
The model-specific settings are:
\begin{itemize}
    \item FLUX.1-dev: 25 inference steps, with TokenRepel applied during the first 3 steps;
    repulsion is applied in all 57 layers, and density preservation is applied in the first 30 layers; $\bar{\theta}=45^\circ$.
    \item FLUX.2-Klein-9B: 4 inference steps, with TokenRepel applied during the first step;
    repulsion is applied in all 32 layers, and density preservation is applied in the first 20 layers; $\bar{\theta}=60^\circ$.
    \item SDXL: 50 inference steps, with TokenRepel applied during the first 5 steps;
    repulsion is applied in all U-Net layers, and density preservation is applied in the downsample and middle layers; $\bar{\theta}=45^\circ$.
\end{itemize}

\paragraph{Image editing.}
For \cref{tab:kontextbench}, we use FLUX.1-Kontext-dev with 28 inference steps and generate four images for each image--instruction pair.
Fidelity metrics are computed against the LLM-generated target prompts,
and diversity metrics are computed among the four edits of each pair.
Token density preservation is applied in all layers.

\paragraph{Hyperparameter analysis.}
For \cref{fig:analysis}, we use FLUX.2-Klein-9B without token density preservation.
We sweep $\delta\in\{10^\circ,20^\circ,30^\circ,40^\circ\}$ with $\bar{\theta}=60^\circ$,
and $\bar{\theta}\in\{20^\circ,40^\circ,60^\circ,80^\circ\}$ with $\delta=30^\circ$.

\paragraph{Token density preservation ablation.}
For \cref{fig:geneval_density_constraint_comparison}, we keep $\bar{\theta}$ fixed and sweep $\delta$:
\begin{itemize}
    \item FLUX.1-dev: $5^\circ, 10^\circ, 15^\circ, 20^\circ$ for w/o P; $10^\circ, 20^\circ, 40^\circ, 60^\circ$ for B; $5^\circ, 10^\circ, 20^\circ, 30^\circ$ for L.
    \item FLUX.2-Klein-9B: $10^\circ, 20^\circ, 30^\circ, 40^\circ$ for w/o P; $20^\circ, 40^\circ, 60^\circ, 80^\circ$ for B and L.
    \item SDXL: $3^\circ, 6^\circ, 9^\circ, 12^\circ$ for w/o P; $5^\circ, 10^\circ, 15^\circ, 20^\circ$ for B and L.
\end{itemize}

\paragraph{Method comparison.}
\Cref{tab:geneval_swap_settings} lists the swept values and fixed settings of each method in \cref{fig:geneval_swap_comparison},
and the settings selected for \cref{tab:comparison}.
The fixed settings mainly follow the original papers.
When a method was not evaluated on a model in its original paper,
we run a small-scale hyperparameter sweep to ensure reasonable performance.

\begin{table}[t]
\centering
\scriptsize
\setlength{\tabcolsep}{4pt}
\caption{Hyperparameter settings used in \cref{fig:geneval_swap_comparison}.
Each method sweeps one parameter and keeps the others fixed.
Bold values are the settings selected for \cref{tab:comparison}.
For TokenRepel, $\delta$ and $\bar{\theta}$ are in degrees.}
\label{tab:geneval_swap_settings}
\begin{tabularx}{\linewidth}{@{}l l l X@{}}
\toprule
Method & Swept parameter & Swept values & Fixed settings \\
\midrule
\multicolumn{4}{@{}l}{\textbf{FLUX.1-dev}} \\
Ours & $\delta$ & $5, 10, \mathbf{20}, 30$ & $\bar{\theta}=45$; steps 1--3; constraint in the first 30 layers \\
NegToMe & $\alpha$ & $\mathbf{0.5}, 0.75, 1, 1.45$ & threshold $0.65$; steps 1--7; all layers \\
PG & coefficient & $\mathbf{0.05}, 0.1, 0.14, 0.19$ & adaptive median bandwidth; every step \\
CADS & noise scale & $\mathbf{0.0005}, 0.001, 0.0013, 0.0016, 0.0022$ & $\tau_1=0.6$, $\tau_2=0.9$, $\psi=1$, rescaling \\
Contextual & $\eta$ & $10^7, 10^8, 10^9, \mathbf{10^{10}}$ & 50 iterations; first step \\
Shield & radius & $0.8, 0.85, 0.9, \mathbf{0.95}, 1$ & $\lambda=0.1$; every step \\
CNO & learning rate & $\mathbf{0.01}, 0.013, 0.017, 0.02$ & 3 optimization steps, $\tau=0.1$, $\gamma=1$, pool size 16 \\
Group & pool size $M$ & $16, \mathbf{32}, 64$ & keep ratio $0.5$, diversity weight $1.5$ \\
\midrule
\multicolumn{4}{@{}l}{\textbf{FLUX.2-Klein-9B}} \\
Ours & $\delta$ & $20, \mathbf{40}, 60, 80$ & $\bar{\theta}=60$; step 1; constraint in the first 20 layers \\
NegToMe & $\alpha$ & $0.5, \mathbf{0.75}, 1, 1.45$ & threshold $0.65$; step 1; all layers \\
PG & coefficient & $0.2, \mathbf{0.3}, 0.4, 0.49$ & adaptive median bandwidth; every step \\
CADS & noise scale & $0.012, \mathbf{0.022}, 0.032, 0.043$ & $\tau_1=0.6$, $\tau_2=0.9$, $\psi=1$, rescaling \\
Contextual & $\eta$ & $2.5\!\times\!10^8, 5\!\times\!10^8, \mathbf{7.5\!\times\!10^8}, 10^9$ & 100 iterations; first step \\
Shield & radius & $1.1, 1.27, \mathbf{1.44}, 1.61$ & $\lambda=0.1$; every step \\
CNO & learning rate & $0.05, 0.09, \mathbf{0.13}, 0.18$ & 3 optimization steps, $\tau=0.1$, $\gamma=1$, pool size 16 \\
Group & pool size $M$ & $8, 16, 32, \mathbf{64}$ & keep ratio $0.1$, diversity weight $1.5$ \\
\midrule
\multicolumn{4}{@{}l}{\textbf{SDXL}} \\
Ours & $\delta$ & $5, \mathbf{10}, 15, 20$ & $\bar{\theta}=45$; steps 1--5; constraint after the down and middle layers \\
NegToMe & $\alpha$ & $0.07, \mathbf{0.14}, 0.21, 0.28$ & threshold $0.7$; steps 1--5; all self-attention layers \\
SPARKE & guidance scale & $0.034, 0.069, \mathbf{0.1}, 0.14$ & $\sigma=0.8$; guidance frequency 10 \\
PG & coefficient & $0.001, 0.002, \mathbf{0.003}, 0.004$ & adaptive median bandwidth; every step \\
CADS & noise scale & $0.005, 0.01, \mathbf{0.015}, 0.02$ & $\tau_1=0.6$, $\tau_2=0.9$, $\psi=1$, rescaling \\
Shield & radius & $0.9, 0.95, 1, \mathbf{1.05}$ & $\lambda=0.1$; every step \\
CNO & learning rate & $0.17, 0.34, \mathbf{0.51}, 0.68$ & 3 optimization steps, $\tau=0.1$, $\gamma=1$, pool size 16 \\
Group & pool size $M$ & $8, 16, \mathbf{32}, 64$ & pruning at every step, diversity weight $1.5$ \\
\bottomrule
\end{tabularx}
\end{table}

\subsection{Differences from the Camera-Ready Version}
\label{app:camera_ready_diff}
In this arXiv version, we provide more complete hyperparameter sweeps and a wider diversity--quality trade-off range for each method on GenEval.
We also run all the ablations and analysis experiments on the larger GenEval benchmark (instead of DrawBench) to obtain more convincing results.
Due to our limited computational resources, the setup differs from the camera-ready version as follows:
\begin{itemize}
    \item FLUX.1-dev uses 25 inference steps instead of 35.
    \item All three models generate images at $512\times512$, whereas the camera-ready version uses $1024\times1024$ for SDXL.
    \item We generate 4 images per prompt instead of 6.
    \item Token density preservation is applied to a different number of layers,
    which gives a better trade-off and faster inference.
    \item We parameterize TokenRepel by the angles $\delta$ and $\bar{\theta}$ in degrees,
    instead of $\alpha=\tan(\delta)$ and $\beta=\cos(\bar{\theta})$,
    since angles directly describe the rotation and the similarity range on the unit hypersphere and are thus easier to interpret.
\end{itemize}
These differences do not change the experimental conclusions.

\end{document}